\documentclass{article} 
\usepackage{iclr2027_conference,times}

\usepackage{amsmath,amsfonts,bm}

\def\eqref#1{equation~\ref{#1}}

\def\1{\bm{1}}

\DeclareMathAlphabet{\mathsfit}{\encodingdefault}{\sfdefault}{m}{sl}
\SetMathAlphabet{\mathsfit}{bold}{\encodingdefault}{\sfdefault}{bx}{n}

\usepackage{float}
\usepackage{hyperref}
\hypersetup{hidelinks,
  pdftitle={LongPuzzleBench: Evaluating GUI Agents on Long-Horizon Visual Puzzles},
  pdfauthor={Bingo Zhang, Haochuan Lu, Zongjie Li, Genjian Li, Ari Yu Zhang, Chaozheng Wang}}
\usepackage{etoc} 
\usepackage{url}
\usepackage{booktabs}
\usepackage{graphicx}
\usepackage{array}
\usepackage{longtable}
\usepackage{xcolor}
\usepackage[font=small,skip=4pt]{caption}
\usepackage{enumitem}
\definecolor{codeC}{HTML}{0072B2}
\definecolor{nativeC}{HTML}{E69F00}
\newcommand{\swatch}[1]{\textcolor{#1}{\rule[0.05ex]{0.95ex}{0.95ex}}\hspace{0.4em}}
\definecolor{findC}{HTML}{1F5F8B}
\newcommand{\finding}[1]{\par\vspace{2pt}\noindent\colorbox{findC!7}{\parbox{\dimexpr\linewidth-2\fboxsep\relax}{\looseness=-1\textbf{#1}\par}}\par\vspace{2pt}}

\title{LongPuzzleBench: Evaluating GUI Agents \\ on Long-Horizon Visual Puzzles}

\author{\normalfont
Bingo Zhang\textsuperscript{1}, Haochuan Lu\textsuperscript{2}, Zongjie Li\textsuperscript{1,3}, Genjian Li\textsuperscript{1}, Ari Yu Zhang\textsuperscript{4}, Chaozheng Wang\textsuperscript{1,5} \\[5pt]
\small \textsuperscript{1}Vera Praxis \\
\small \textsuperscript{2}Tencent \\
\small \textsuperscript{3}The Hong Kong University of Science and Technology \\
\small \textsuperscript{4}Independent Researcher \\
\small \textsuperscript{5}Department of Computer Science and Engineering, The Chinese University of Hong Kong
}

\iclrfinalcopy
\begin{document}
\etocdepthtag.toc{mtmain}

\maketitle
\lhead{Preprint}

\begin{abstract}
GUI agents need long-horizon visual reasoning: they must interpret a changing interface while keeping a multi-step plan viable as earlier actions constrain later ones. Existing benchmarks evaluate grounding, computer use, and game play, but rarely test whether agents stay coherent across long chains of coupled decisions. Long-horizon visual puzzles expose this capability directly: a legal move that looks like progress can make the puzzle unsolvable, and the loss shows only several moves later. We introduce LongPuzzleBench, 114 levels in six puzzle games played through native GUI actions, where one objective can take a human over a thousand actions on persistent boards and dead ends go unannounced. With Native GUI Actions alone, the strongest agents solve most objectives, but success falls sharply on harder, longer boards: seven of ten general-purpose agents solve nothing harder than Medium, and none completes Bolt Unscrew Hard, which a human solves along with every other objective. Code Execution CUA does not close this gap, and its scores mix visual solving with algorithmic search. Controlled diagnostics trace these failures to one limitation that neither rules, state hints, nor failure memory removes: agents judge each move by the visible progress it makes, not by the future options it leaves.
\par\medskip
\noindent\textbf{Code: }\href{https://github.com/AzureStarz/LongPuzzleBench}{\textcolor{codeC}{\textbf{\texttt{github.com/AzureStarz/LongPuzzleBench}}}}
\end{abstract}

\vspace{-5pt}
\section{Introduction}
\vspace{-5pt}
\label{sec:introduction}
\looseness=-1 GUI agents act by reading a rendered interface and choosing the next input \citep{pmlr-v235-zheng24e,hong2024cogagent,qin2025uitarspioneeringautomatedgui}. When a task spans many decisions and earlier actions change what later actions can achieve, reading the current screen is not enough: the agent must track how the state has changed, anticipate what each action rules out, and revise its plan when an option is lost. We call this capability \emph{long-horizon visual reasoning}.

\begin{figure}[!t]
\centering
\includegraphics[width=\linewidth]{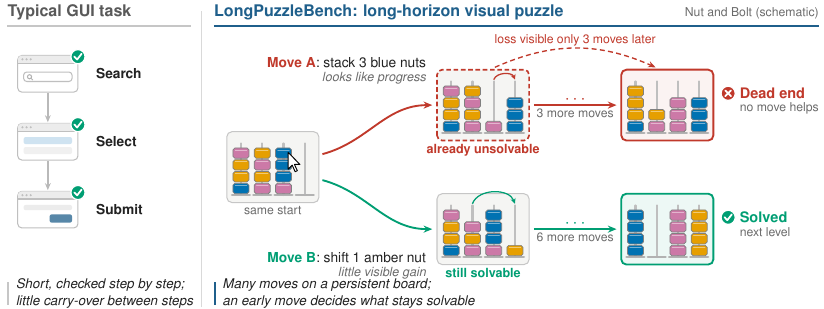}
\caption{Early moves decide later solvability. In a typical GUI task (left), each step is checked at once and has little effect on later steps. In a long-horizon visual puzzle (right), two legal moves from the same board diverge: Move~A looks like progress but already makes the level unsolvable, which shows only three moves later, whereas Move~B keeps it solvable. The Nut and Bolt board is schematic; both outcomes are verified by exhaustive search.}
\label{fig:puzzle-reasoning}
\end{figure}

\looseness=-1 Existing benchmarks measure parts of this capability but rarely isolate it. GUI benchmarks test grounding \citep{cheng-etal-2024-seeclick} and multi-step computer use \citep{koh-etal-2024-visualwebarena,NEURIPS2024_0b82662b,Xie_2024,ICLR2025_01a83bc2}, visual puzzle benchmarks test reasoning over a single static configuration \citep{chia-etal-2024-puzzlevqa,ghosal-etal-2025-algopuzzlevqa,zhang2025puzzlebenchfullydynamicevaluation}, and game benchmarks test planning and rule discovery, often through predefined commands or with resets left to the harness \citep{ICLR2025_f0b1515b,zhang2026videogamebenchvisionlanguagemodelscomplete,foundation2026arcagi3newchallengefrontier}. A long trajectory alone does not imply a long reasoning horizon: a GUI task can take many steps that are each checked at once (Figure~\ref{fig:puzzle-reasoning}, left), and a long real-time game episode mixes planning with perception, timing, and control. Neither isolates whether agents judge each move by the options it leaves.

\looseness=-1 Long-horizon visual puzzles supply this demand. Their horizon is set by decision dependency, not elapsed time or action count: rules tie the current board to the options left after each move, so a legal move that looks like progress can make the puzzle unsolvable \citep{junghanns2001sokoban}, with the loss surfacing only several moves later (Figure~\ref{fig:puzzle-reasoning}, right). Self-paced play is reasoning-dominant rather than timing-dependent, and reproducible states make each failure replayable.

We introduce \emph{LongPuzzleBench}, 114 levels of long-horizon visual puzzles in six games, grouped into 16 objectives that chain up to ten levels under one time budget. Every move persists on the board, and play is long even for people: a human needs a median of 204 GUI actions per objective and 1,106 for a single Nut and Bolt level (Figure~\ref{fig:benchmark-overview}). When a move makes a level unsolvable, the game never says so; later moves simply stop working. Agents play under \emph{self-directed gameplay}, acting only through the rendered interface, so each failure traces back to their own decisions.

We evaluate agents in two settings: \emph{Native GUI Actions}, which admits only coordinate-based GUI input, and \emph{Code Execution CUA}, in which a computer-use agent (CUA) may also write and run code. With Native GUI Actions, the strongest agents solve most objectives (GPT-6-Astra 91.7\% OSR), but success falls sharply on harder, longer boards: seven of ten general-purpose agents solve no objective harder than Medium, the four GUI-specialized models complete none, and no agent completes Bolt Unscrew Hard, whereas a human solves all 16 objectives. Code Execution CUA does not close this gap: it raises OSR for six of ten agents, not in proportion to their use of code, and its scores mix visual solving with algorithmic search.

\looseness=-1 Tracing failing trajectories and intervening on reproduced states (Section~\ref{sec:analysis}), we find that agents spend the free capacity that keeps a puzzle solvable and cycle through earlier states when no move shows visible progress. At a dead end, they mostly read rejected moves as missed clicks rather than as a lost board. Told that the board is lost, they restart, but most rebuild a dead end, with or without memory of the failed attempt. Stating the rules raises rule knowledge far more than completion. These results point to one limitation: agents judge each move by the visible progress it makes, not by the future options it leaves.

Our contributions are:
\begin{itemize}[leftmargin=1.5em,topsep=2pt,itemsep=1pt,parsep=0pt]
    \item LongPuzzleBench, long-horizon visual puzzles played through native GUI actions, with reproducible states and a validated refresh pipeline (Section~\ref{sec:benchmark}Arxiv).
    \item An evaluation of 14 agents in two action settings against a human reference: the strongest agents solve most objectives, but success falls sharply on harder, longer boards, and Code Execution CUA does not close the gap (Section~\ref{sec:experiments}).
    \item Controlled interventions on reproduced states that trace failures to judging moves by visible progress rather than preserved options, a limitation that rules, state hints, and failure memory do not remove (Section~\ref{sec:analysis}).
\end{itemize}

\vspace{-5pt}
\section{Related Work}
\vspace{-5pt}
\label{sec:related-work}
\label{sec:benchmark-comparison}
\looseness=-1 \noindent\textbf{GUI and computer-use benchmarks.} GUI benchmarks evaluate multi-step task completion through rendered interfaces, from web navigation \citep{NEURIPS2023_5950bf29,ICLR2024_4410c071,koh-etal-2024-visualwebarena} and enterprise workflows \citep{pmlr-v235-drouin24a,NEURIPS2024_0b82662b} to desktop and mobile environments with executable success checks \citep{Xie_2024,ICLR2025_01a83bc2}. Agents either prompt general-purpose multimodal models \citep{pmlr-v235-zheng24e} or train GUI-specialized models for screen grounding and action \citep{cheng-etal-2024-seeclick,hong2024cogagent,qin2025uitarspioneeringautomatedgui}. Their steps are checkable against an instructed goal, and no legal action is designed to make the goal unreachable. LongPuzzleBench instead requires reasoning about what a persistent board still allows.

\begin{table}[t]
\centering
\scriptsize
\setlength{\tabcolsep}{2pt}
\renewcommand{\arraystretch}{1.08}
\caption{Structural comparison with adjacent benchmarks. $\checkmark$:~yes; $\times$:~no; $\dagger$:~in a subset of tasks or one track; --:~not applicable or not reported. Property definitions and per-benchmark notes: Appendix~\ref{sec:comparison-criteria}.}
\label{tab:benchmark-comparison}
\begin{tabular*}{\linewidth}{@{\extracolsep{\fill}}llccccc@{}}
\toprule
& & \multicolumn{3}{c}{What solving requires} & \multicolumn{2}{c}{How the agent must act} \\
\cmidrule(lr){3-5}\cmidrule(lr){6-7}
Benchmark & Environment & \shortstack{Visual\\puzzle state} & \shortstack{Delayed\\consequences} & \shortstack{Dead-end\\states} & \shortstack{Agent-owned\\recovery} & \shortstack{Native\\GUI actions} \\
\midrule
PuzzleBench \citep{zhang2025puzzlebenchfullydynamicevaluation} & Puzzle VQA & $\checkmark$ & $\times$ & -- & -- & -- \\
\addlinespace[2pt]
VisualWebArena \citep{koh-etal-2024-visualwebarena} & Web apps & $\times$ & $\times$ & $\times$ & -- & $\times$ \\
AndroidWorld \citep{ICLR2025_01a83bc2} & Mobile apps & $\times$ & $\times$ & $\times$ & -- & $\checkmark$ \\
\addlinespace[2pt]
BALROG \citep{ICLR2025_f0b1515b} & Text/grid games & $\checkmark^{\dagger}$ & $\checkmark$ & $\checkmark$ & $\times$ & $\times$ \\
VideoGameBench \citep{zhang2026videogamebenchvisionlanguagemodelscomplete} & Video games & $\checkmark^{\dagger}$ & $\checkmark$ & $\checkmark$ & $\times$ & $\checkmark$ \\
GameWorld \citep{ouyang2026gameworldstandardizedverifiableevaluation} & Browser games & $\checkmark^{\dagger}$ & $\checkmark$ & $\checkmark$ & $\times$ & $\checkmark^{\dagger}$ \\
\addlinespace[2pt]
VLATIM \citep{triebel2026visionlanguagemodelshumanlikelogicalproblemsolving} & Physics puzzles & $\checkmark$ & $\times$ & $\times$ & $\times$ & $\checkmark^{\dagger}$ \\
ARC-AGI-3 \citep{foundation2026arcagi3newchallengefrontier} & Grid games & $\checkmark$ & $\checkmark$ & $\checkmark$ & $\checkmark$ & $\times$ \\
\midrule
\textbf{LongPuzzleBench} & Board puzzles & $\checkmark$ & $\checkmark$ & $\checkmark^{\dagger}$ & $\checkmark$ & $\checkmark$ \\
\bottomrule
\end{tabular*}
\end{table}

\looseness=-1 \noindent\textbf{Reasoning in puzzles and games.} Visual puzzle benchmarks pose questions about a static puzzle image that no action changes \citep{chia-etal-2024-puzzlevqa,ghosal-etal-2025-algopuzzlevqa,zhang2025puzzlebenchfullydynamicevaluation}. Interactive puzzle benchmarks require agents to act on their interpretations: VLATIM evaluates physics-puzzle manipulation in single-attempt episodes \citep{triebel2026visionlanguagemodelshumanlikelogicalproblemsolving}, and ARC-AGI-3, the closest to our setting, tests rule and goal discovery with agent-initiated resets but acts through abstract game actions rather than a rendered interface \citep{foundation2026arcagi3newchallengefrontier}. Game benchmarks add delayed consequences over extended trajectories: SmartPlay and BALROG evaluate planning across game suites \citep{ICLR2024_063d1fad,ICLR2025_f0b1515b}, VideoGameBench and GameWorld test rendered-frame play, including in real time \citep{zhang2026videogamebenchvisionlanguagemodelscomplete,ouyang2026gameworldstandardizedverifiableevaluation}, LMGame-Bench diagnoses game playing through modular agent components \citep{ICLR2026_83a4ea71}, and EscapeBench studies implicit goals in text-based escape rooms \citep{qian-etal-2025-escapebench}. These suites contain dead ends, but the harness typically resets failed episodes, and evaluation targets general game play rather than self-paced puzzles whose early moves silently decide solvability. In text-based planning, models given a domain's full rules still produce invalid plans \citep{NEURIPS2023_7a92bcde}, and explicit lookahead search improves planning and puzzle solving \citep{NEURIPS2023_271db992,hao-etal-2023-reasoning}. Studies of recovery examine verbal feedback across trials \citep{shinn2023reflexion}, intrinsic self-correction \citep{ICLR2024_8b4add8b}, and policy-induced GUI errors \citep{bu2026recoveringpolicyinducederrorsbenchmarking}, but not failures that must be inferred from accumulated visual consequences. LongPuzzleBench combines all five properties in Table~\ref{tab:benchmark-comparison}, and its dead ends are unannounced: the agent must infer from the board that progress is impossible.

\vspace{-5pt}
\section{LongPuzzleBench}
\begin{figure}[t]
\centering
\includegraphics[width=\linewidth]{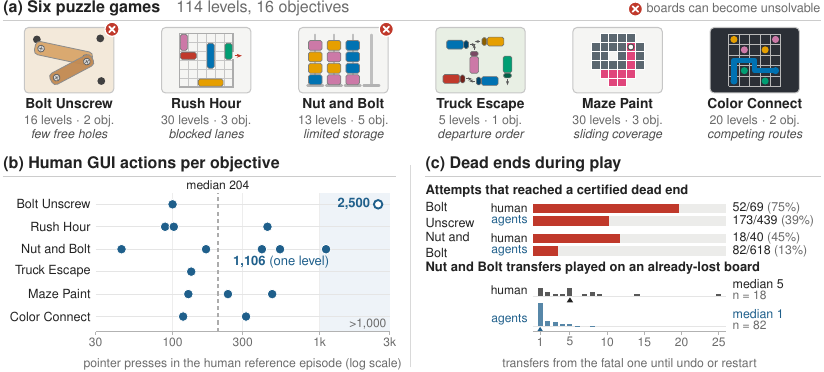}
\caption{LongPuzzleBench at benchmark scale. (a) Six games; red marks those whose boards can become unsolvable. (b) Pointer presses in the human reference episode for each of the 16 objectives, all solved; dashed line: median. (c) Level attempts that reached a certified dead end, and the Nut and Bolt transfers played on an already-lost board, counting the fatal one, before an undo or restart. Agent counts cover the 384 evaluation episodes of Table~\ref{tab:main-results}.}
\label{fig:benchmark-overview}
\end{figure}

\vspace{-5pt}
\label{sec:benchmark}
LongPuzzleBench tests how GUI agents preserve future options on persistent boards. It comprises 114 levels in six puzzle games, grouped into 16 game--difficulty \emph{objectives} (Figure~\ref{fig:benchmark-overview}).

\suppressfloats[t]

\label{sec:environments}
\looseness=-1 \noindent\textbf{Moves that constrain later options.} Each game pairs a rendered board with rules under which a legal move can use up something that later moves need: free holes in Bolt Unscrew, empty storage in Nut and Bolt, open lanes in Rush Hour, shared space in Color Connect, and a workable departure order in Truck Escape (Figure~\ref{fig:benchmark-overview}a). Maze Paint instead requires full coverage under sliding movement, so the last cells can be reached only through moves that paint nothing new. In Bolt Unscrew and Nut and Bolt, such a move can make the level unsolvable without any warning: later moves simply stop working. These dead ends are common in real play: of the human's 69 Bolt Unscrew level attempts, 52 reached a certified deadlock, as did 173 of 439 agent attempts; on Nut and Bolt, the human played a median of five transfers on an already-unsolvable board before undoing or restarting (Figure~\ref{fig:benchmark-overview}c). In the other four games, a poor move cannot make the level unsolvable but can lengthen the path to completion. The games span clicks, drags, and swipes and one to five difficulty tiers (Appendix~\ref{sec:catalogue-details}).

\noindent\textbf{Long horizon at two scales.} Moves create causal dependencies by changing the options available to later moves. The \emph{within-level causal horizon} is how long these constraints persist; action counts describe interaction length. Nut and Bolt Nightmare, a single level, took the human 1,106 pointer presses. Across levels, 15 of the 16 objectives chain three to ten target levels under one budget; an unsolved level blocks advancement. The human used a median of 204 presses per objective, and 14 of the 16 needed at least 100 (Figure~\ref{fig:benchmark-overview}b). The first scale tests planning within a board; the second tests sustained solving across boards, carrying mechanics forward and deciding when to restart.

\label{sec:task-definition}
\label{sec:instrumentation}
\looseness=-1 \noindent\textbf{Evaluation protocol.} An objective $o=(g,d,L_o)$ specifies a game $g$, a difficulty $d$, and target levels $L_o$. The agent receives the GUI action specification and a single instruction, ``Complete all levels of [game] at [difficulty] difficulty in the Game Suite,'' with no manual, solution, or navigation procedure. From the suite home screen, it reads screenshots and uses clicks, drags, key presses, and waits to enter the game, solve each level, advance, and restart through the game's own controls (Appendix Figure~\ref{fig:self-directed-gameplay}), all within one 60-minute budget. This self-directed setting applies one interface to every game and leaves decisions such as abandoning a lost board to the agent. An automatic reset at a dead end would remove the need to infer failure from the board. Agent-owned recovery tests this inference together with the ability to revise a plan, while failed attempts consume the same budget as successful play. Board state and grading signals stay private to the evaluator, and any intermediate board state can be restored exactly, which Section~\ref{sec:analysis} uses as checkpoints for interventions.

\label{sec:construction-validation}
\noindent\textbf{Construction and validation.} Games and levels are reproduced from screenshots or other visual references with a coding model and run in a browser suite with deterministic launch parameters. Before release, GUI agents playtest each candidate to expose defects, which are repaired or excluded \citep{huang2026guiagentscontinualgame}; solvers verify solutions where implemented, including all 30 Maze Paint levels; and manual play confirms that every released level is solvable. Validated levels and games extend the catalogue between snapshots to limit repeated exposure to fixed tasks \citep{ICLR2025_94074dd5}, and each comparison fixes the catalogue, runtime, and seed (Appendix~\ref{sec:construction-details}).

\vspace{-5pt}
\section{How Well Do Agents Solve Long-Horizon Visual Puzzles?}
\vspace{-5pt}
\label{sec:experiments}
We evaluate general-purpose and GUI-specialized agents on all 16 objectives against a human reference. Because an agent that can run code can transcribe a board and search for moves offline, we measure performance in two settings that differ in the action channel.

\vspace{-3pt}
\subsection{Setup}
\label{sec:eval-settings}
\label{sec:common-protocol}
\noindent\textbf{Settings.} Both settings follow the protocol of Section~\ref{sec:task-definition} and share objectives, budget, metrics, and a dedicated Chromium browser at $1280\times900$. With \emph{Native GUI Actions}, the agent sees screenshots and acts only through clicks, drags, key presses, and other coordinate-based inputs; it cannot execute programs of its own, so this setting measures visual solving without computational assistance. With \emph{Code Execution CUA}, the agent may also write and run code over screenshots and the data it derives from them, while every change to the game must still pass through GUI inputs; how it uses computation is analyzed from its trajectory. Game source, hidden state, evaluator outputs, and external retrieval are prohibited in both settings (Appendix Table~\ref{tab:setting-interfaces}).

\looseness=-1 \noindent\textbf{Agents.} Ten general-purpose backbones (six proprietary, four open-weight) run in the Codex scaffold in both settings; with Native GUI Actions, the agent may call only a single GUI-action tool. Four GUI-specialized models \citep{qin2025uitarspioneeringautomatedgui,bai2025qwen3vltechnicalreport}, trained to map screenshots to coordinate actions, run with Native GUI Actions in an autonomous harness that exposes the same action set in each model's own output format. A human player provides one reference episode per objective (Appendix~\ref{sec:settings-protocol}).

\looseness=-1 \noindent\textbf{Metrics.} The primary metric, \emph{objective success rate} (OSR), is the percentage of objectives whose target levels are all solved within an episode, averaged over objectives within each game and then equally over the six games; all completion metrics use this game-macro weighting. \emph{Level completion} (LC) is the percentage of target levels solved, and \emph{fine-grained progress} (FGP) also credits the largest fraction of the goal reached on the level where an episode stops (e.g., cells painted or pairs connected). LC$_\mathrm{hard}$ restricts LC to tiers above each game's easiest. The \emph{level pass rate} (LPR) is the share of reached levels solved, the \emph{50\% level horizon} (H$_{50}$; cf.\ \citealp{NEURIPS2025_85069585}) is the number of consecutive levels that at least half of the episodes clear on the ten-level objectives, and the \emph{premature exit rate} (PER) counts episodes the agent abandons. Under Code Execution CUA, the \emph{native GUI action ratio} (NGAR) is the share of steps that consist only of screen observation or native GUI input (Appendix~\ref{sec:metrics}).

\begin{table}[t]
\centering\footnotesize
\setlength{\tabcolsep}{1.6pt}
\caption{Main results. Swatch colors match Figure~\ref{fig:capability-profile}; $^{\diamond}$open-weight. Bold: best within each setting. Values in \%; H$_{50}$ in levels (10+: at least half the episodes clear all ten). --: undefined or not applicable.}
\label{tab:main-results}
\begin{tabular}{@{}>{\raggedright\arraybackslash}p{0.335\linewidth}*{8}{>{\centering\arraybackslash}p{0.0705\linewidth}}@{}}
\toprule
& \multicolumn{3}{c}{Completion} & \multicolumn{2}{c}{Horizon} & {\scriptsize Difficulty} & \multicolumn{2}{c}{Behavior} \\
\cmidrule(lr){2-4}\cmidrule(lr){5-6}\cmidrule(lr){7-7}\cmidrule(lr){8-9}
Agent & OSR $\uparrow$ & LC $\uparrow$ & FGP $\uparrow$ & LPR $\uparrow$ & H$_{50}$ $\uparrow$ & {\scriptsize LC$_\mathrm{hard}$\,$\uparrow$} & PER $\downarrow$ & {\scriptsize NGAR\,$\uparrow$} \\
\midrule
\textit{Human reference} & 100.0 & 100.0 & 100.0 & 100.0 & 10+ & 100.0 & 0.0 & -- \\
\addlinespace[3pt]
\multicolumn{9}{@{}l}{\swatch{nativeC}\textbf{Native GUI Actions}\enspace{\scriptsize\textit{general-purpose, Codex}}} \\
\hspace{0.6em}GPT-6-Astra & \textbf{91.7} & \textbf{94.8} & \textbf{94.8} & \textbf{97.9} & 10+ & \textbf{87.5} & 0.0 & -- \\
\hspace{0.6em}GPT-6-Sol & 82.8 & 91.0 & 92.2 & 95.4 & 10+ & 80.3 & 0.0 & -- \\
\hspace{0.6em}GPT-6-Luna & 33.9 & 40.0 & 43.8 & 46.7 & 2 & 9.0 & 68.8 & -- \\
\hspace{0.6em}GPT-5.6-Sol & 71.1 & 84.3 & 86.0 & 91.7 & 10+ & 64.7 & 6.2 & -- \\
\hspace{0.6em}GPT-5.6-Terra & 48.3 & 61.3 & 64.7 & 78.1 & 5 & 32.5 & 56.2 & -- \\
\hspace{0.6em}GPT-5.6-Luna & 45.0 & 53.3 & 56.4 & 60.3 & 6 & 16.0 & 62.5 & -- \\
\hspace{0.6em}DeepSeek 4.1 Flash$^{\diamond}$ & 53.9 & 72.0 & 74.9 & 77.5 & 10+ & 48.0 & 12.5 & -- \\
\hspace{0.6em}Qwen3.8 Flash$^{\diamond}$ & 53.9 & 67.2 & 71.5 & 76.0 & 10+ & 37.0 & 0.0 & -- \\
\hspace{0.6em}GLM-5.3-Flash$^{\diamond}$ & 33.9 & 51.9 & 55.2 & 60.6 & 5 & 12.7 & 12.5 & -- \\
\hspace{0.6em}Kimi-k3$^{\diamond}$ & 56.7 & 64.4 & 67.2 & 73.3 & 10+ & 30.7 & 0.0 & -- \\
\addlinespace[2pt]
\multicolumn{9}{@{}l}{\hspace{1.35em}{\scriptsize\textit{GUI-specialized, autonomous harness}}} \\
\hspace{0.6em}UI-TARS-1.5-7B$^{\diamond}$ & 0.0 & 5.5 & 6.6 & -- & 0 & 0.0 & 87.5 & -- \\
\hspace{0.6em}Qwen3-VL-32B-Thinking$^{\diamond}$ & 0.0 & 1.1 & 6.4 & -- & 0 & 0.0 & 0.0 & -- \\
\hspace{0.6em}Qwen3-VL-30B-A3B-Thinking$^{\diamond}$ & 0.0 & 1.0 & 2.5 & -- & 0 & 0.0 & 18.8 & -- \\
\hspace{0.6em}Qwen3-VL-8B-Thinking$^{\diamond}$ & 0.0 & 2.2 & 3.4 & -- & 0 & 0.0 & 0.0 & -- \\
\addlinespace[3pt]
\multicolumn{9}{@{}l}{\swatch{codeC}\textbf{Code Execution CUA}\enspace{\scriptsize\textit{general-purpose, Codex}}} \\
\hspace{0.6em}GPT-6-Astra & \textbf{91.7} & \textbf{95.8} & \textbf{96.7} & \textbf{98.3} & 10+ & \textbf{90.0} & 0.0 & 65.7 \\
\hspace{0.6em}GPT-6-Sol & 80.0 & 82.2 & 83.6 & 82.2 & 10+ & 58.3 & 18.8 & 41.1 \\
\hspace{0.6em}GPT-6-Luna & 36.1 & 57.5 & 61.2 & 71.0 & 6 & 32.3 & 75.0 & 65.4 \\
\hspace{0.6em}GPT-5.6-Sol & 88.3 & 91.6 & 91.9 & 94.7 & 10+ & 80.8 & 0.0 & 8.5 \\
\hspace{0.6em}GPT-5.6-Terra & 45.0 & 68.7 & 72.7 & 75.0 & 10+ & 48.7 & 62.5 & 33.1 \\
\hspace{0.6em}GPT-5.6-Luna & 39.4 & 45.6 & 49.6 & 50.2 & 4 & 10.7 & 68.8 & 52.5 \\
\addlinespace[2pt]
\hspace{0.6em}DeepSeek 4.1 Flash$^{\diamond}$ & 67.2 & 76.6 & 79.2 & 82.9 & 10+ & 56.8 & 6.2 & 11.7 \\
\hspace{0.6em}Qwen3.8 Flash$^{\diamond}$ & 76.1 & 81.0 & 82.6 & 86.6 & 10+ & 64.2 & 0.0 & 17.5 \\
\hspace{0.6em}GLM-5.3-Flash$^{\diamond}$ & 39.4 & 52.2 & 57.1 & 72.3 & 3 & 21.2 & 0.0 & 51.0 \\
\hspace{0.6em}Kimi-k3$^{\diamond}$ & 70.6 & 77.6 & 78.7 & 85.0 & 10+ & 59.0 & 0.0 & 59.3 \\
\bottomrule
\end{tabular}
\end{table}

\vspace{-3pt}
\subsection{Native GUI play falls short on harder tiers}
\label{sec:main-results}
\label{sec:backbone-evaluation}
With Native GUI Actions, most agents solve mainly the easiest puzzles, whereas the human solves all 16 objectives (Table~\ref{tab:main-results}). Of the 84 successful episodes of the general-purpose backbones, 52 fall on each game's easiest tier, 20 on Medium, and 12 on harder tiers, all by GPT-6-Astra (6), GPT-6-Sol (4), and GPT-5.6-Sol (2). The other seven solve 70.0--100.0\% of easiest-tier levels but only 9.0--48.0\% of harder-tier levels, and GPT-6-Astra solves every objective except Bolt Unscrew Hard. These successes suggest that operating the interface is not the main barrier; the barrier is carrying solving to boards that demand longer chains of coupled moves. GUI-specialized models do not close this gap: the four clear eight levels in 64 episodes, complete no objective, and reach an FGP of only 2.5--6.6, so screenshot-grounding training does not carry over to sustained puzzle play.

\vspace{-3pt}
\subsection{Code Execution CUA mixes visual solving with search}
\label{sec:native-gui-analysis}
\looseness=-1 Under Code Execution CUA, OSR ranges from 36.1\% to 91.7\% (Table~\ref{tab:main-results}), but agents reach these scores in two ways. GPT-6-Astra acts natively in 65.7\% of steps and reaches the same OSR as with Native GUI Actions (91.7\%). GPT-5.6-Sol, Qwen3.8 Flash, and DeepSeek 4.1 Flash, second, fourth, and sixth in OSR, do so in only 8.5\%, 17.5\%, and 11.7\% of steps; their other steps search for moves over transcribed boards, parse screenshots pixel by pixel, or query page geometry. None accesses game source or evaluator state (Appendix~\ref{sec:settings-protocol}), so these steps are permitted, but, as in program-aided reasoning \citep{pmlr-v202-gao23f}, they move the solving outside the model's visual reasoning. An agent that transcribes a board can search it offline and replay the solution in one call; on levels solved in both settings, such play takes 60\% of the time needed with Native GUI Actions. The saving raises completion only where play with Native GUI Actions is still solving levels at the deadline, as in the 11 timeouts with Native GUI Actions that Code Execution CUA completes (Appendix~\ref{sec:setting-mechanism}).

Removing code lowers OSR for six of the ten backbones, but not in proportion to their reliance on computation. DeepSeek 4.1 Flash, with an NGAR of 11.7\%, falls from 67.2\% to 53.9\% OSR, and Kimi-k3, with an NGAR of 59.3\%, falls by a similar margin, from 70.6\% to 56.7\% (Appendix~\ref{sec:settings-results}). GPT-6-Sol, GPT-5.6-Terra, and GPT-5.6-Luna even score higher with Native GUI Actions (by 2.8, 3.3, and 5.6 points), partly because each abandons more episodes under Code Execution CUA. The settings also differ in tool integration (Appendix~\ref{sec:settings-protocol}), so these gaps characterize each setting as a whole rather than the action channel alone. We therefore treat Native GUI Actions results as the measure of visual solving and read every Code Execution CUA score together with its NGAR.

\vspace{-3pt}
\subsection{Where performance breaks down}
\label{sec:difficulty-horizon}
\begin{figure}[t]
\centering
\includegraphics[width=\linewidth,trim=0 2pt 0 9pt,clip]{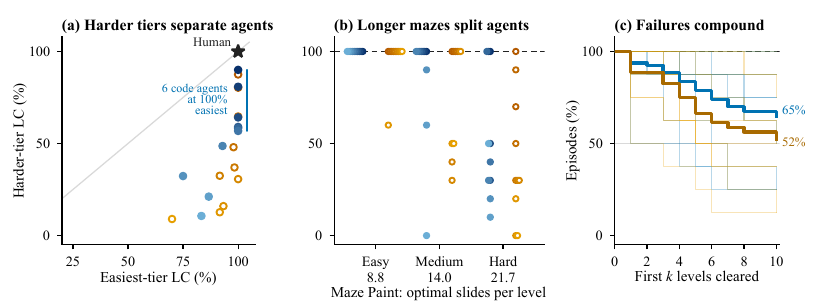}
\caption{Performance falls as the required horizon grows (all ten general-purpose agents in both settings; values as in Table~\ref{tab:main-results}). Filled blue: Code Execution CUA; hollow orange: Native GUI Actions; darker shades: higher LC$_\mathrm{hard}$; star and dashed lines: human. (a) Level completion on each game's easiest tier against its harder tiers. (b) Maze Paint level completion per tier, labeled with mean solver-optimal slides per level. (c) Episodes that clear the first $k$ levels of the eight ten-level objectives; thin: agents, bold: mean over agents.}
\label{fig:capability-profile}
\end{figure}

\noindent\textbf{Harder tiers separate agents that tie on easy ones.} Under Code Execution CUA, six agents solve every easiest-tier level, yet their LC$_\mathrm{hard}$ spans 56.8--90.0\% (Figure~\ref{fig:capability-profile}a). Maze Paint ties this spread to solution length: solver-optimal solutions grow from 8.8 to 21.7 slides per level from Easy to Hard, all ten agents clear every Easy level under Code Execution CUA, and Hard-level LC ranges from 10\% to 100\% (Figure~\ref{fig:capability-profile}b). Easy tiers saturate; the harder tiers measure long-horizon solving.

\noindent\textbf{Small per-level gaps become large objective gaps.} Levels are won in catalogue order in every valid episode, so one unsolved level forfeits all later ones. With Native GUI Actions, GPT-6-Sol exceeds GPT-5.6-Sol by 3.7 points in LPR but by 11.7 points in OSR, and Kimi-k3 passes 92.6\% of the levels it reaches on the ten-level objectives, yet only four of its eight episodes clear all ten. Across agents, only 65\% of episodes under Code Execution CUA and 52\% with Native GUI Actions clear all ten levels (Figure~\ref{fig:capability-profile}c), so scoring levels in isolation would understate how far agents remain from sustained solving.

\noindent\textbf{Some agents stop early.} GPT-5.6-Luna, GPT-6-Luna, and GPT-5.6-Terra, the three weakest proprietary agents in OSR, voluntarily end 62.5--75.0\% of their episodes under Code Execution CUA and 56.2--68.8\% with Native GUI Actions; since progression stays with the agent, the decision to keep playing is part of the measured capability.

\vspace{-3pt}
\subsection{Humans and agents succeed on different puzzles}
\label{sec:capability-diagnostics}
\label{sec:human-agent-diagnostic}
No episode in either setting clears more than four of the eight Bolt Unscrew Hard levels, all of which the human completes, yet GPT-6-Astra solves Nut and Bolt far more directly than the human. We compare the human with Astra's evaluation episodes at shared puzzle states, mainly those with Native GUI Actions, which match the human's interface (Appendix~\ref{sec:human-agent-protocol}). From the same Nut and Bolt Nightmare state, Astra needs 65 stack-state changes (67 under Code Execution CUA) and never breaks a solved stack, whereas the human needs 431 changes and 17 resets; Astra also uses fewer moves on all six Maze Paint and Rush Hour objectives, though more gestures in Color Connect. Bolt Unscrew Hard reverses the ranking. On level~4, the human and both Astra episodes first fill the four free holes and deadlock before removing a board. The human then changes the opening, removes 12 boards on the next attempt, and clears the level on the third; Astra under Code Execution CUA repeats the zero-board dead end in its first five attempts and clears the level on the eighth, and Astra with Native GUI Actions reaches the level with two minutes left (Appendix Figure~\ref{fig:human-astra}). The reversal follows what each puzzle demands. Nut and Bolt rewards bookkeeping over a visible state in which progress accumulates as solved stacks; Bolt Unscrew rewards preserving capacity, and a free hole is not visible progress. Current agents thus match or exceed the human at maintaining explicit state but fall short when a move must be judged by the options it preserves. This contrast rests on one human trajectory per setting; Section~\ref{sec:rq1} tests the capacity mechanism across agents and retries.

\vspace{-5pt}
\section{Failure Diagnosis: Why Do Agents Fail?}
\vspace{-5pt}
\label{sec:analysis}
The gap lies on harder tiers and long level sequences (Section~\ref{sec:experiments}). With Native GUI Actions, 66--80\% of each general-purpose agent's missing level completion lies in levels it never reached (Appendix~\ref{sec:rq1-observed}), so an objective is lost inside the few levels where progress stops. This section follows such a level from the move that loses it to the attempt that follows: how agents lose a solvable board (RQ1), whether they notice (RQ2), whether recovery and memory produce a better attempt (RQ3), and whether knowing the rules would have prevented the loss (RQ4). Each diagnostic restores a reproducible puzzle state and varies one factor while the agent, task, and starting board stay fixed (Appendix~\ref{sec:diagnostic-common}). Mechanism evidence comes mainly from Bolt Unscrew and Maze Paint, where the hidden cost of a move can be counted. Controlled studies use GPT-5.6-Sol and repeat the key contrasts with GPT-5.6-Terra and GPT-5.6-Luna; in this section, Sol, Terra, and Luna denote these three GPT-5.6 models.

\begin{figure}[t]
\centering
\includegraphics[width=\linewidth,trim=0 4pt 0 6pt,clip]{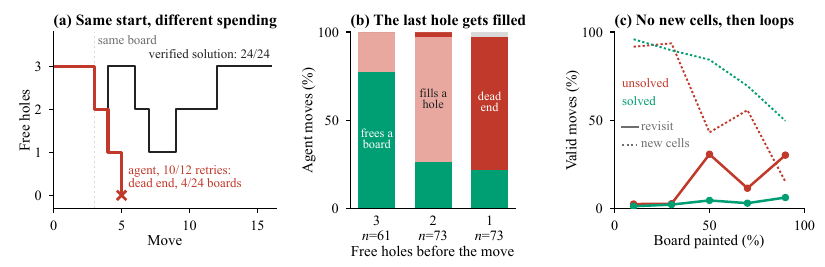}
\caption{Moves that look like progress spend solvability; moves without visible progress loop (RQ1). (a) Free holes along Bolt Unscrew Hard level~1 from one start: a verified solution and the agent path repeated in 10 of 12 retries. (b) Effect of agent moves by free holes before the move, over all Bolt Unscrew retries. (c) Maze Paint, 260 attempts with Native GUI Actions: share of valid moves that revisit an exact earlier state or paint new cells, by board coverage.}
\label{fig:section5-solvability}
\end{figure}

\vspace{-3pt}
\subsection{Moves that look like progress spend solvability}
\label{sec:rq1}
\label{sec:progression-analysis}
\noindent\textbf{RQ1: How do agents lose a solvable board?} Two games make the hidden cost of a move measurable. In Bolt Unscrew, a screw moved without freeing a board occupies one of a few free holes, and a level with no free hole left is a dead end. In Maze Paint, the last cells can be reached only by sliding across painted ones, so late moves rarely show progress.

\finding{Finding 1. Agents choose moves by what they change now, rarely by the options they leave.}

\noindent\textbf{Visible progress spends capacity.} From the same start of Bolt Unscrew Hard level~1, the agent and a verified solution both remove four boards in three moves and keep two free holes. The solution's next move frees three more boards and reopens a hole. The agent instead fills the remaining holes and deadlocks with 20 of 24 boards still attached, along the same path in 10 of 12 retries (Figure~\ref{fig:section5-solvability}a). The pattern holds across retries: with one free hole left, three quarters of agent moves fill it and end the level at once (Figure~\ref{fig:section5-solvability}b), as do 15 of 17 first moves from solvable level starts with a single free hole. Many are misjudgments, not slips: in 28 of the 55 fatal moves, the agent's stated reason predicts that the move will release a board.

\looseness=-1 \noindent\textbf{Without visible progress, agents loop.} Early in a Maze Paint level, nearly every move paints new cells. As the board fills, solved attempts slow down but keep reaching new states. Unsolved attempts cycle instead: in the last fifth of the board, only 15\% of their valid moves paint new cells and 30\% return to an exact earlier state, against 50\% and 6\% in solved attempts (Figure~\ref{fig:section5-solvability}c). This contrast is descriptive, since solved and unsolved attempts are not matched for difficulty; per-agent rates appear in Table~\ref{tab:rq1-maze}. Replacing one costly move with a shortest-path move from the same state barely changes completion (3/27 versus 4/27), which suggests that the loss accumulates over many moves rather than arising from one error.

Both games point to the same gap. Agents respond to what the screen shows, such as empty holes and unpainted cells, but not to the quantities that decide whether the level can still be finished: the capacity a move consumes and the states already tried. Neither is drawn on the board, so a more accurate reading of the current screen is unlikely to repair the failure on its own.

\vspace{-3pt}
\subsection{Dead ends look like missed clicks}
\label{sec:rq2}
\noindent\textbf{RQ2: Do agents notice that a board is lost?} We restore 24 certified Bolt Unscrew dead ends and let Sol continue with no hint (A), with a truthful statement that the board requires recovery (B), or with B plus the name of the restart control (C). Solvable boards serve as controls.

\finding{Finding 2. Agents tend to blame a lost board on their own execution, not on the puzzle state.}

\begin{figure}[t]
\centering
\includegraphics[width=\linewidth,trim=0 4pt 0 5pt,clip]{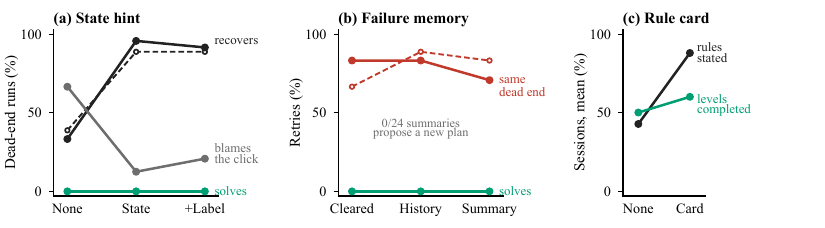}
\caption{Each support moves the behavior it targets far more than it moves completion (RQ2--RQ4). (a) Sol on 24 certified Bolt Unscrew dead ends with no hint, state information, or state information plus the restart label. (b) Retries from the restarted board with the prior attempt cleared, kept as history, or replaced by an agent-written summary. (c) Sol sessions without and with the rule card (P0), means over 16 objectives. Solid: GPT-5.6-Sol; dashed: GPT-5.6-Sol, -Terra, and -Luna (18 units per condition).}
\label{fig:section5-supports}
\end{figure}

On a dead board, every move is rejected, just as a mis-aimed click would be. Without a hint, two thirds of runs explain the rejection as an execution error (``the screw did not move; try the other hole'') and retry other targets. Only 8 of 24 runs recover, and only one restarts promptly (Figure~\ref{fig:section5-supports}a). The diagnosis is within reach: 7 of those 8 runs state the cause correctly (``both storage holes are occupied without releasing a plank''). One sentence of state information reverses the attribution. Click-error explanations fall to 3 of 24 runs and recovery rises to 23 of 24, and from 7 to 16 of 18 across three models, while no solvable control triggers a restart. Naming the restart control adds nothing, and in Nut and Bolt no model recovers under any condition (Appendix~\ref{sec:rq2-diagnostics}).

\looseness=-1 In Bolt Unscrew, the bottleneck is therefore the inference behind the restart, not the restart action: whether an unexpected outcome is attributed to the world or to one's own execution. Blaming execution is a reasonable default in GUI control, where a rejected action often means a missed click \citep{cheng-etal-2024-seeclick,Xie_2024}, but in a puzzle that never announces a dead end, it leaves most dead ends unrecognized.

\vspace{-3pt}
\subsection{Recovery resets the board, not the plan}
\label{sec:rq3}
\looseness=-1 \noindent\textbf{RQ3: Does recovery lead to a better attempt?} A restart returns the agent to a solvable board, and the restarted agent can keep direct evidence of how its last attempt failed. We examine the recovered RQ2 runs and 72 matched retries that branch each agent-owned Bolt Unscrew failure from the same restarted board. The prior attempt is cleared (M0), kept as native history (M1), or replaced by an agent-written failure summary \citep{shinn2023reflexion} of equal size (M2).

\finding{Finding 3. Agents remember what they tried, not why it failed, so memory does not keep them from rebuilding the same dead end.}

None of the 153 dead-end continuations in RQ2 completes the level. Of the 45 runs that recover under B or C, 34 build a new certified dead end, at a median of 129 seconds after recovery, and 29 of these make visible progress first. Memory does not change completion: no retry completes the level under any memory condition (0/126, including the three-model repeat; Figure~\ref{fig:section5-supports}b). The memory variants choose different first moves in 20 of 24 sources, yet 79\% of retries end in the exact dead end of their source, and agents holding the full history return to it at least as often as agents without it. The summaries suggest why. All 24 attribute the failure to execution or missing observations (``the click did not register''), none identifies that the free holes ran out, and none proposes a different plan; 15 instead direct the next attempt to resume the failed one (Figure~\ref{fig:section5-support-details}b).

A failure record helps only if it encodes a causal lesson that constrains the next plan. These records encode actions and open questions: the attribution that hides a dead end (Finding~2) also keeps it out of memory, so a restart restores the capacity without a reason to spend it differently.

\vspace{-3pt}
\subsection{Knowing the rules is not planning with them}
\label{sec:rq4}
\noindent\textbf{RQ4: Would knowing the rules have prevented these failures?} Findings~1--3 could reflect rules the agent never inferred, and full sessions add a second burden: entering games and advancing levels. RQ4 crosses a mechanics card that states each game's rules (K1 versus K0) with a deterministic controller for entry and advancement (P1 versus P0) on all 16 objectives, leaving every puzzle move to the agent. Periodic probes measure which rules the agent can state.

\finding{Finding 4. Rule knowledge helps, but agents can state a puzzle's rules well before they can plan with them.}

Agents do discover rules through play: without the card, most rules that Sol ever states correctly are first stated during play (Appendix~\ref{sec:mechanism-analyses}). The card doubles rule knowledge, from 43\% to 88\% of rules stated correctly, yet raises level completion only from 50\% to 60\% (Figure~\ref{fig:section5-supports}c), and 7 of the 16 sessions whose final probe states every rule correctly still leave levels unsolved. Across three models, the card adds about nine LC points on average, with large variation across objectives; workflow control removes 19--29 model actions per session but does not reliably add levels (Figure~\ref{fig:section5-support-details}a).

Rules describe what one move does; solving requires predicting what a sequence of moves does to future options. Knowing that a screw needs an empty hole does not tell an agent that filling the last one ends the level, the misjudgment behind the dead ends of RQ1. Before committing to the move, the agent must predict whether it will release a board and reopen capacity for subsequent moves.

\vspace{-5pt}
\section{Conclusion}
\vspace{-5pt}
\label{sec:conclusion}
LongPuzzleBench evaluates long-horizon visual reasoning through self-paced GUI puzzles in which early decisions constrain later moves and agents must recognize and recover from dead ends. Performance falls on harder, longer boards, and computational assistance does not close the gap. Controlled diagnostics trace failures to a common limitation: agents judge moves by immediate visible progress while overlooking the future options they consume. State hints, failure memory, and rule cards change the targeted behaviors far more than completion. These findings make preserving future options a central challenge for visual puzzle agents. Evaluating this capability requires tasks where decisions remain consequential over many moves and success depends on keeping a plan viable through completion.

\subsection*{AI use statement}
AI tools were used only to polish the language, grammar, and phrasing of this paper. They were not used to generate research ideas, experimental conclusions, or scientific findings. The authors take full responsibility for the content of the manuscript.

\subsection*{Ethics statement}
This work evaluates AI agents on puzzle games in a controlled, locally run browser environment. It does not involve human subjects, sensitive personal data, or real-world high-risk deployment; the human reference consists only of in-game actions and board states. We therefore identify no ethical concerns that require further discussion.

\subsection*{Reproducibility statement}
Sections~\ref{sec:benchmark} and~\ref{sec:eval-settings} summarize the benchmark, the two action settings, and the evaluation protocol. Appendices~\ref{sec:catalogue-details} and~\ref{sec:construction-details} specify the games, objectives, progress potentials, and deterministic launch parameters; Appendices~\ref{sec:full-setup}--\ref{sec:compute-cost} give the episode protocol, agent endpoints and inference settings, metric definitions, and token usage; and Appendix~\ref{sec:diagnostic-common} gives the shared design of the controlled diagnostics. Each comparison fixes the catalogue, runtime, and seed. The benchmark code, game suite, and evaluation harness are available at \url{https://anonymous.4open.science/r/LongPuzzleBench-CD1A}.


\bibliography{iclr2027_conference}

@inproceedings{ICLR2024_4410c071,
 author = {Zhou, Shuyan and Xu, Frank F and Zhu, Hao and Zhou, Xuhui and Lo, Robert and Sridhar, Abishek and Cheng, Xianyi and Ou, Tianyue and Bisk, Yonatan and Fried, Daniel and Alon, Uri and Neubig, Graham},
 booktitle = {International Conference on Learning Representations},
 editor = {B. Kim and Y. Yue and S. Chaudhuri and K. Fragkiadaki and M. Khan and Y. Sun},
 pages = {15585--15606},
 title = {{WebArena: A Realistic Web Environment for Building Autonomous Agents}},
 url = {https://proceedings.iclr.cc/paper_files/paper/2024/file/4410c0711e9154a7a2d26f9b3816d1ef-Paper-Conference.pdf},
 volume = {2024},
 year = {2024}
}

@inproceedings{koh-etal-2024-visualwebarena,
    title = "{{V}isual{W}eb{A}rena: Evaluating Multimodal Agents on Realistic Visual Web Tasks}",
    author = "Koh, Jing Yu  and
      Lo, Robert  and
      Jang, Lawrence  and
      Duvvur, Vikram  and
      Lim, Ming  and
      Huang, Po-Yu  and
      Neubig, Graham  and
      Zhou, Shuyan  and
      Salakhutdinov, Russ  and
      Fried, Daniel",
    editor = "Ku, Lun-Wei  and
      Martins, Andre  and
      Srikumar, Vivek",
    booktitle = "Proceedings of the 62nd Annual Meeting of the Association for Computational Linguistics (Volume 1: Long Papers)",
    month = aug,
    year = "2024",
    address = "Bangkok, Thailand",
    publisher = "Association for Computational Linguistics",
    url = "https://aclanthology.org/2024.acl-long.50/",
    doi = "10.18653/v1/2024.acl-long.50",
    pages = "881--905"
}

@InProceedings{pmlr-v235-drouin24a,
  title = 	 {{W}ork{A}rena: How Capable are Web Agents at Solving Common Knowledge Work Tasks?},
  author =       {Drouin, Alexandre and Gasse, Maxime and Caccia, Massimo and Laradji, Issam H. and Del Verme, Manuel and Marty, Tom and Vazquez, David and Chapados, Nicolas and Lacoste, Alexandre},
  booktitle = 	 {Proceedings of the 41st International Conference on Machine Learning},
  pages = 	 {11642--11662},
  year = 	 {2024},
  editor = 	 {Salakhutdinov, Ruslan and Kolter, Zico and Heller, Katherine and Weller, Adrian and Oliver, Nuria and Scarlett, Jonathan and Berkenkamp, Felix},
  volume = 	 {235},
  series = 	 {Proceedings of Machine Learning Research},
  month = 	 {21--27 Jul},
  publisher =    {PMLR},
  url = 	 {https://proceedings.mlr.press/v235/drouin24a.html},
}

@inproceedings{NEURIPS2024_0b82662b,
 author = {Boisvert, L\'{e}o and Thakkar, Megh and Gasse, Maxime and Caccia, Massimo and De Chezelles, Thibault Le Sellier and Cappart, Quentin and Chapados, Nicolas and Lacoste, Alexandre and Drouin, Alexandre},
 booktitle = {Advances in Neural Information Processing Systems},
 doi = {10.52202/079017-0195},
 editor = {A. Globerson and L. Mackey and D. Belgrave and A. Fan and U. Paquet and J. Tomczak and C. Zhang},
 pages = {5996--6051},
 publisher = {Curran Associates, Inc.},
 title = {{WorkArena++: Towards Compositional Planning and Reasoning-based Common Knowledge Work Tasks}},
 url = {https://proceedings.neurips.cc/paper_files/paper/2024/file/0b82662b6c32e887bb252a74d8cb2d5e-Paper-Datasets_and_Benchmarks_Track.pdf},
 volume = {37},
 year = {2024}
}

@inproceedings{ICLR2025_f0b1515b,
 author = {Paglieri, Davide and Cupia{\l}, Bart{\l}omiej and Coward, Samuel and Piterbarg, Ulyana and Wo{\l}czyk, Maciej and Khan, Akbir and Pignatelli, Eduardo and Kuci{\'n}ski, {\L}ukasz and Pinto, Lerrel and Fergus, Rob and Foerster, Jakob and Parker-Holder, Jack and Rockt{\"a}schel, Tim},
 booktitle = {International Conference on Learning Representations},
 editor = {Y. Yue and A. Garg and N. Peng and F. Sha and R. Yu},
 pages = {96666--96702},
 title = {{BALROG: Benchmarking Agentic LLM and VLM Reasoning On Games}},
 url = {https://proceedings.iclr.cc/paper_files/paper/2025/file/f0b1515be276f6ba82b4f2b25e50bef0-Paper-Conference.pdf},
 volume = {2025},
 year = {2025}
}

@inproceedings{ICLR2024_063d1fad,
  author = {Wu, Yue and Tang, Xuan and Mitchell, Tom and Li, Yuanzhi},
  booktitle = {International Conference on Learning Representations},
  editor = {B. Kim and Y. Yue and S. Chaudhuri and K. Fragkiadaki and M. Khan and Y. Sun},
  pages = {1543--1561},
  title = {{SmartPlay: A Benchmark for LLMs as Intelligent Agents}},
  url = {https://proceedings.iclr.cc/paper_files/paper/2024/file/063d1fadbde557466bf5bcffe814a64e-Paper-Conference.pdf},
  volume = {2024},
  year = {2024}
}

@misc{zhang2026videogamebenchvisionlanguagemodelscomplete,
  title = {{VideoGameBench: Can Vision-Language Models Complete Popular Video Games?}},
  author = {Zhang, Alex L. and Griffiths, Thomas L. and Narasimhan, Karthik R. and Press, Ofir},
  year = {2026},
  eprint = {2505.18134},
  archivePrefix = {arXiv},
  primaryClass = {cs.AI},
  url = {https://arxiv.org/abs/2505.18134}
}

@inproceedings{qian-etal-2025-escapebench,
    title = "{{E}scape{B}ench: Towards Advancing Creative Intelligence of Language Model Agents}",
    author = "Qian, Cheng  and
      Han, Peixuan  and
      Luo, Qinyu  and
      He, Bingxiang  and
      Chen, Xiusi  and
      Zhang, Yuji  and
      Du, Hongyi  and
      Yao, Jiarui  and
      Yang, Xiaocheng  and
      Zhang, Denghui  and
      Li, Yunzhu  and
      Ji, Heng",
    editor = "Che, Wanxiang  and
      Nabende, Joyce  and
      Shutova, Ekaterina  and
      Pilehvar, Mohammad Taher",
    booktitle = "Proceedings of the 63rd Annual Meeting of the Association for Computational Linguistics (Volume 1: Long Papers)",
    month = jul,
    year = "2025",
    address = "Vienna, Austria",
    publisher = "Association for Computational Linguistics",
    url = "https://aclanthology.org/2025.acl-long.39/",
    doi = "10.18653/v1/2025.acl-long.39",
    pages = "798--820",
    ISBN = "979-8-89176-251-0"
}

@inproceedings{NEURIPS2025_85069585,
 author = {Kwa, Thomas and West, Ben and Becker, Joel and Deng, Amy and Garcia, Katharyn and Hasin, Max and Jawhar, Sami and Kinniment, Megan and Rush, Nate and Von Arx, Sydney and Bloom, Ryan and Broadley, Thomas and Du, Haoxing and Goodrich, Brian and Jurkovic, Nikola and Miles, Luke and Nix, Seraphina and Lin, Tao and Parikh, Neev and Rein, David and Koba Sato, Lucas Jun and Wijk, Hjalmar and Ziegler, Daniel and Barnes, Elizabeth and Chan, Lawrence},
 booktitle = {Advances in Neural Information Processing Systems},
 doi = {10.52202/085713-3086},
 editor = {D. Belgrave and C. Zhang and H. Lin and R. Pascanu and P. Koniusz and M. Ghassemi and N. Chen},
 pages = {92213--92266},
 publisher = {Curran Associates, Inc.},
 title = {{Measuring AI Ability to Complete Long Software Tasks}},
 url = {https://proceedings.neurips.cc/paper_files/paper/2025/file/85069585133c4c168c865e65d72e9775-Paper-Conference.pdf},
 volume = {38, Main Conference},
 year = {2025}
}

@inproceedings{Xie_2024,
  series = {NeurIPS 2024},
  title = {{OSWorld: Benchmarking Multimodal Agents for Open-Ended Tasks in Real Computer Environments}},
  url = {https://doi.org/10.52202/079017-1650},
  doi = {10.52202/079017-1650},
  booktitle = {Advances in Neural Information Processing Systems 37},
  publisher = {Neural Information Processing Systems Foundation, Inc. (NeurIPS)},
  author = {Xie, Tianbao and Zhang, Danyang and Chen, Jixuan and Li, Xiaochuan and Zhao, Siheng and Cao, Ruisheng and Hua, Toh Jing and Cheng, Zhoujun and Shin, Dongchan and Lei, Fangyu and Liu, Yitao and Xu, Yiheng and Zhou, Shuyan and Savarese, Silvio and Xiong, Caiming and Zhong, Victor and Yu, Tao},
  year = {2024},
  pages = {52040--52094}
}

@inproceedings{ICLR2025_01a83bc2,
  author = {Rawles, Chris and Clinckemaillie, Sarah and Chang, Yifan and Waltz, Jonathan and Lau, Gabrielle and Fair, Marybeth and Li, Alice and Bishop, William and Li, Wei and Campbell-Ajala, Folawiyo and Toyama, Daniel and Berry, Robert and Tyamagundlu, Divya and Lillicrap, Timothy and Riva, Oriana},
  booktitle = {International Conference on Learning Representations},
  editor = {Y. Yue and A. Garg and N. Peng and F. Sha and R. Yu},
  pages = {406--441},
  title = {{AndroidWorld: A Dynamic Benchmarking Environment for Autonomous Agents}},
  url = {https://proceedings.iclr.cc/paper_files/paper/2025/file/01a83bc2f2732a58e6aa731e659e7101-Paper-Conference.pdf},
  volume = {2025},
  year = {2025}
}

@misc{ouyang2026gameworldstandardizedverifiableevaluation,
  title = {{GameWorld: Towards Standardized and Verifiable Evaluation of Multimodal Game Agents}},
  author = {Ouyang, Mingyu and Hu, Siyuan and Lin, Kevin Qinghong and Ng, Hwee Tou and Shou, Mike Zheng},
  year = {2026},
  eprint = {2604.07429},
  archivePrefix = {arXiv},
  primaryClass = {cs.CV},
  url = {https://arxiv.org/abs/2604.07429}
}

@misc{bu2026recoveringpolicyinducederrorsbenchmarking,
  title = {{Recovering Policy-Induced Errors: Benchmarking and Trajectory Synthesis for Robust GUI Agents}},
  author = {Bu, Tianpeng and Liu, Xin and Chen, Qihua and Jiang, Hao and Li, Shurui and Duan, Hongtao and Jiang, Lu and Hu, Lulu and Yang, Bin and Zhang, Minying},
  year = {2026},
  eprint = {2605.29447},
  archivePrefix = {arXiv},
  primaryClass = {cs.CV},
  url = {https://arxiv.org/abs/2605.29447},
  note = {ICML 2026 Spotlight}
}

@inproceedings{shinn2023reflexion,
  title = {{Reflexion: Language Agents with Verbal Reinforcement Learning}},
  author = {Shinn, Noah and Cassano, Federico and Gopinath, Ashwin and Narasimhan, Karthik and Yao, Shunyu},
  booktitle = {Advances in Neural Information Processing Systems},
  volume = {36},
  pages = {8634--8652},
  year = {2023},
  doi = {10.52202/075280-0377},
  url = {https://proceedings.neurips.cc/paper_files/paper/2023/hash/1b44b878bb782e6954cd888628510e90-Abstract-Conference.html}
}

@inproceedings{ICLR2025_94074dd5,
 author = {Jain, Naman and Han, King and Gu, Alex and Li, Wen-Ding and Yan, Fanjia and Zhang, Tianjun and Wang, Sida and Solar-Lezama, Armando and Sen, Koushik and Stoica, Ion},
 booktitle = {International Conference on Learning Representations},
 editor = {Y. Yue and A. Garg and N. Peng and F. Sha and R. Yu},
 pages = {58791--58831},
 title = {{LiveCodeBench: Holistic and Contamination Free Evaluation of Large Language Models for Code}},
 url = {https://proceedings.iclr.cc/paper_files/paper/2025/file/94074dd5a072d28ff75a76dabed43767-Paper-Conference.pdf},
 volume = {2025},
 year = {2025}
}

@misc{huang2026guiagentscontinualgame,
      title={{GUI Agents for Continual Game Generation}}, 
      author={Yixu Huang and Bo Li and Na Li and Zhe Wang and Kaijie Chen and Haonan Ge and Qingyi Si and Yuanzhe Shen and Ruihan Yang and Guangjing Wang and Hongcheng Guo},
      year={2026},
      eprint={2605.28258},
      archivePrefix={arXiv},
      primaryClass={cs.SE},
      url={https://arxiv.org/abs/2605.28258}, 
}

@misc{foundation2026arcagi3newchallengefrontier,
      title={{ARC-AGI-3: A New Challenge for Frontier Agentic Intelligence}}, 
      author={{ARC Prize Foundation}},
      year={2026},
      eprint={2603.24621},
      archivePrefix={arXiv},
      primaryClass={cs.AI},
      url={https://arxiv.org/abs/2603.24621}, 
}

@misc{ying2026aigamestorescalableopenended,
      title={{AI Gamestore: Scalable, Open-Ended Evaluation of Machine General Intelligence with Human Games}}, 
      author={Lance Ying and Ryan Truong and Prafull Sharma and Kaiya Ivy Zhao and Nathan Cloos and Kelsey R. Allen and Thomas L. Griffiths and Katherine M. Collins and José Hernández-Orallo and Phillip Isola and Samuel J. Gershman and Joshua B. Tenenbaum},
      year={2026},
      eprint={2602.17594},
      archivePrefix={arXiv},
      primaryClass={cs.AI},
      url={https://arxiv.org/abs/2602.17594}, 
}

@misc{triebel2026visionlanguagemodelshumanlikelogicalproblemsolving,
      title={{Do Vision-Language-Models show human-like logical problem-solving capability in point and click puzzle games?}}, 
      author={Maximilian Triebel and Marco Menner and Dominik Helfenstein},
      year={2026},
      eprint={2605.11223},
      archivePrefix={arXiv},
      primaryClass={cs.AI},
      url={https://arxiv.org/abs/2605.11223}, 
}

@misc{zhang2025puzzlebenchfullydynamicevaluation,
      title={{PuzzleBench: A Fully Dynamic Evaluation Framework for Large Multimodal Models on Puzzle Solving}}, 
      author={Zeyu Zhang and Zijian Chen and Zicheng Zhang and Yuze Sun and Yuan Tian and Ziheng Jia and Chunyi Li and Xiaohong Liu and Xiongkuo Min and Guangtao Zhai},
      year={2025},
      eprint={2504.10885},
      archivePrefix={arXiv},
      primaryClass={cs.CV},
      url={https://arxiv.org/abs/2504.10885}, 
}

@inproceedings{ICLR2026_83a4ea71,
 author = {Hu, Lanxiang and Huo, Mingjia and Zhang, Yuxuan and Yu, Haoyang and Xing, Eric P and Stoica, Ion and Rosing, Tajana and Jin, Haojian and Zhang, Hao},
 booktitle = {International Conference on Learning Representations},
 editor = {C. Vondrick and B. Hariharan and C. Raffel and L. Pinto and D. Yang and A. Faust},
 pages = {81308--81356},
 title = {{lmgame-Bench: How Good are LLMs at Playing Games?}},
 url = {https://proceedings.iclr.cc/paper_files/paper/2026/file/83a4ea71b13bc86308a2bd0b5e07fb61-Paper-Conference.pdf},
 volume = {2026},
 year = {2026}
}

@inproceedings{cheng-etal-2024-seeclick,
    title = "{S}ee{C}lick: Harnessing {GUI} Grounding for Advanced Visual {GUI} Agents",
    author = "Cheng, Kanzhi  and
      Sun, Qiushi  and
      Chu, Yougang  and
      Xu, Fangzhi  and
      Li, Yantao  and
      Zhang, Jianbing  and
      Wu, Zhiyong",
    editor = "Ku, Lun-Wei  and
      Martins, Andre  and
      Srikumar, Vivek",
    booktitle = "Proceedings of the 62nd Annual Meeting of the Association for Computational Linguistics (Volume 1: Long Papers)",
    month = aug,
    year = "2024",
    address = "Bangkok, Thailand",
    publisher = "Association for Computational Linguistics",
    url = "https://aclanthology.org/2024.acl-long.505/",
    doi = "10.18653/v1/2024.acl-long.505",
    pages = "9313--9332"
}

@inproceedings{NEURIPS2023_5950bf29,
 author = {Deng, Xiang and Gu, Yu and Zheng, Boyuan and Chen, Shijie and Stevens, Sam and Wang, Boshi and Sun, Huan and Su, Yu},
 booktitle = {Advances in Neural Information Processing Systems},
 doi = {10.52202/075280-1220},
 editor = {A. Oh and T. Naumann and A. Globerson and K. Saenko and M. Hardt and S. Levine},
 pages = {28091--28114},
 publisher = {Curran Associates, Inc.},
 title = {{Mind2Web: Towards a Generalist Agent for the Web}},
 url = {https://proceedings.neurips.cc/paper_files/paper/2023/file/5950bf290a1570ea401bf98882128160-Paper-Datasets_and_Benchmarks.pdf},
 volume = {36},
 year = {2023}
}

@InProceedings{pmlr-v235-zheng24e,
  title = 	 {{GPT}-4{V}(ision) is a Generalist Web Agent, if Grounded},
  author =       {Zheng, Boyuan and Gou, Boyu and Kil, Jihyung and Sun, Huan and Su, Yu},
  booktitle = 	 {Proceedings of the 41st International Conference on Machine Learning},
  pages = 	 {61349--61385},
  year = 	 {2024},
  editor = 	 {Salakhutdinov, Ruslan and Kolter, Zico and Heller, Katherine and Weller, Adrian and Oliver, Nuria and Scarlett, Jonathan and Berkenkamp, Felix},
  volume = 	 {235},
  series = 	 {Proceedings of Machine Learning Research},
  month = 	 {21--27 Jul},
  publisher =    {PMLR},
  url = 	 {https://proceedings.mlr.press/v235/zheng24e.html}
}

@inproceedings{hong2024cogagent,
  title = {{CogAgent: A Visual Language Model for GUI Agents}},
  author = {Hong, Wenyi and Wang, Weihan and Lv, Qingsong and Xu, Jiazheng and Yu, Wenmeng and Ji, Junhui and Wang, Yan and Wang, Zihan and Dong, Yuxiao and Ding, Ming and Tang, Jie},
  booktitle = {Proceedings of the IEEE/CVF Conference on Computer Vision and Pattern Recognition (CVPR)},
  publisher = {IEEE},
  pages = {14281--14290},
  year = {2024},
  doi = {10.1109/CVPR52733.2024.01354}
}

@misc{qin2025uitarspioneeringautomatedgui,
  title = {{UI-TARS: Pioneering Automated GUI Interaction with Native Agents}},
  author = {Qin, Yujia and Ye, Yining and Fang, Junjie and Wang, Haoming and Liang, Shihao and Tian, Shizuo and Zhang, Junda and Li, Jiahao and Li, Yunxin and Huang, Shijue and Zhong, Wanjun and Li, Kuanye and Yang, Jiale and Miao, Yu and Lin, Woyu and Liu, Longxiang and Jiang, Xu and Ma, Qianli and Li, Jingyu and Xiao, Xiaojun and Cai, Kai and Li, Chuang and Zheng, Yaowei and Jin, Chaolin and Li, Chen and Zhou, Xiao and Wang, Minchao and Chen, Haoli and Li, Zhaojian and Yang, Haihua and Liu, Haifeng and Lin, Feng and Peng, Tao and Liu, Xin and Shi, Guang},
  year = {2025},
  eprint = {2501.12326},
  archivePrefix = {arXiv},
  primaryClass = {cs.AI},
  url = {https://arxiv.org/abs/2501.12326}
}

@misc{bai2025qwen3vltechnicalreport,
  title = {{Qwen3-VL Technical Report}},
  author = {Bai, Shuai and Cai, Yuxuan and Chen, Ruizhe and Chen, Keqin and Chen, Xionghui and Cheng, Zesen and Deng, Lianghao and Ding, Wei and Gao, Chang and Ge, Chunjiang and Ge, Wenbin and Guo, Zhifang and Huang, Qidong and Huang, Jie and Huang, Fei and Hui, Binyuan and Jiang, Shutong and Li, Zhaohai and Li, Mingsheng and Li, Mei and Li, Kaixin and Lin, Zicheng and Lin, Junyang and Liu, Xuejing and Liu, Jiawei and Liu, Chenglong and Liu, Yang and Liu, Dayiheng and Liu, Shixuan and Lu, Dunjie and Luo, Ruilin and Lv, Chenxu and Men, Rui and Meng, Lingchen and Ren, Xuancheng and Ren, Xingzhang and Song, Sibo and Sun, Yuchong and Tang, Jun and Tu, Jianhong and Wan, Jianqiang and Wang, Peng and Wang, Pengfei and Wang, Qiuyue and Wang, Yuxuan and Xie, Tianbao and Xu, Yiheng and Xu, Haiyang and Xu, Jin and Yang, Zhibo and Yang, Mingkun and Yang, Jianxin and Yang, An and Yu, Bowen and Zhang, Fei and Zhang, Hang and Zhang, Xi and Zheng, Bo and Zhong, Humen and Zhou, Jingren and Zhou, Fan and Zhou, Jing and Zhu, Yuanzhi and Zhu, Ke},
  year = {2025},
  eprint = {2511.21631},
  archivePrefix = {arXiv},
  primaryClass = {cs.CV},
  url = {https://arxiv.org/abs/2511.21631}
}

@inproceedings{chia-etal-2024-puzzlevqa,
    title = "{P}uzzle{VQA}: Diagnosing Multimodal Reasoning Challenges of Language Models with Abstract Visual Patterns",
    author = "Chia, Yew Ken  and
      Toh, Vernon  and
      Ghosal, Deepanway  and
      Bing, Lidong  and
      Poria, Soujanya",
    editor = "Ku, Lun-Wei  and
      Martins, Andre  and
      Srikumar, Vivek",
    booktitle = "Findings of the Association for Computational Linguistics: ACL 2024",
    month = aug,
    year = "2024",
    address = "Bangkok, Thailand",
    publisher = "Association for Computational Linguistics",
    url = "https://aclanthology.org/2024.findings-acl.962/",
    doi = "10.18653/v1/2024.findings-acl.962",
    pages = "16259--16273"
}

@inproceedings{ghosal-etal-2025-algopuzzlevqa,
    title = "{A}lgo{P}uzzle{VQA}: Diagnosing Multimodal Reasoning Challenges of Language Models with Algorithmic Multimodal Puzzles",
    author = "Ghosal, Deepanway  and
      Toh, Vernon  and
      Chia, Yew Ken  and
      Poria, Soujanya",
    editor = "Chiruzzo, Luis  and
      Ritter, Alan  and
      Wang, Lu",
    booktitle = "Proceedings of the 2025 Conference of the Nations of the Americas Chapter of the Association for Computational Linguistics: Human Language Technologies (Volume 1: Long Papers)",
    month = apr,
    year = "2025",
    address = "Albuquerque, New Mexico",
    publisher = "Association for Computational Linguistics",
    url = "https://aclanthology.org/2025.naacl-long.486/",
    doi = "10.18653/v1/2025.naacl-long.486",
    pages = "9615--9632"
}

@article{junghanns2001sokoban,
  title = {{Sokoban: Enhancing General Single-Agent Search Methods Using Domain Knowledge}},
  author = {Junghanns, Andreas and Schaeffer, Jonathan},
  journal = {Artificial Intelligence},
  volume = {129},
  number = {1--2},
  pages = {219--251},
  year = {2001},
  publisher = {Elsevier},
  doi = {10.1016/S0004-3702(01)00109-6}
}

@inproceedings{NEURIPS2023_7a92bcde,
 author = {Valmeekam, Karthik and Marquez, Matthew and Olmo, Alberto and Sreedharan, Sarath and Kambhampati, Subbarao},
 booktitle = {Advances in Neural Information Processing Systems},
 doi = {10.52202/075280-1693},
 editor = {A. Oh and T. Naumann and A. Globerson and K. Saenko and M. Hardt and S. Levine},
 pages = {38975--38987},
 publisher = {Curran Associates, Inc.},
 title = {{PlanBench: An Extensible Benchmark for Evaluating Large Language Models on Planning and Reasoning about Change}},
 url = {https://proceedings.neurips.cc/paper_files/paper/2023/file/7a92bcdede88c7afd108072faf5485c8-Paper-Datasets_and_Benchmarks.pdf},
 volume = {36},
 year = {2023}
}

@inproceedings{NEURIPS2023_271db992,
 author = {Yao, Shunyu and Yu, Dian and Zhao, Jeffrey and Shafran, Izhak and Griffiths, Tom and Cao, Yuan and Narasimhan, Karthik},
 booktitle = {Advances in Neural Information Processing Systems},
 doi = {10.52202/075280-0517},
 editor = {A. Oh and T. Naumann and A. Globerson and K. Saenko and M. Hardt and S. Levine},
 pages = {11809--11822},
 publisher = {Curran Associates, Inc.},
 title = {{Tree of Thoughts: Deliberate Problem Solving with Large Language Models}},
 url = {https://proceedings.neurips.cc/paper_files/paper/2023/file/271db9922b8d1f4dd7aaef84ed5ac703-Paper-Conference.pdf},
 volume = {36},
 year = {2023}
}

@inproceedings{hao-etal-2023-reasoning,
    title = "Reasoning with Language Model is Planning with World Model",
    author = "Hao, Shibo  and
      Gu, Yi  and
      Ma, Haodi  and
      Hong, Joshua  and
      Wang, Zhen  and
      Wang, Daisy  and
      Hu, Zhiting",
    editor = "Bouamor, Houda  and
      Pino, Juan  and
      Bali, Kalika",
    booktitle = "Proceedings of the 2023 Conference on Empirical Methods in Natural Language Processing",
    month = dec,
    year = "2023",
    address = "Singapore",
    publisher = "Association for Computational Linguistics",
    url = "https://aclanthology.org/2023.emnlp-main.507/",
    doi = "10.18653/v1/2023.emnlp-main.507",
    pages = "8154--8173"
}

@inproceedings{ICLR2024_8b4add8b,
 author = {Huang, Jie and Chen, Xinyun and Mishra, Swaroop and Zheng, Huaixiu Steven and Yu, Adams Wei and Song, Xinying and Zhou, Denny},
 booktitle = {International Conference on Learning Representations},
 editor = {B. Kim and Y. Yue and S. Chaudhuri and K. Fragkiadaki and M. Khan and Y. Sun},
 pages = {32808--32824},
 title = {{Large Language Models Cannot Self-Correct Reasoning Yet}},
 url = {https://proceedings.iclr.cc/paper_files/paper/2024/file/8b4add8b0aa8749d80a34ca5d941c355-Paper-Conference.pdf},
 volume = {2024},
 year = {2024}
}

@InProceedings{pmlr-v202-gao23f,
  title = 	 {{PAL}: Program-aided Language Models},
  author =       {Gao, Luyu and Madaan, Aman and Zhou, Shuyan and Alon, Uri and Liu, Pengfei and Yang, Yiming and Callan, Jamie and Neubig, Graham},
  booktitle = 	 {Proceedings of the 40th International Conference on Machine Learning},
  pages = 	 {10764--10799},
  year = 	 {2023},
  editor = 	 {Krause, Andreas and Brunskill, Emma and Cho, Kyunghyun and Engelhardt, Barbara and Sabato, Sivan and Scarlett, Jonathan},
  volume = 	 {202},
  series = 	 {Proceedings of Machine Learning Research},
  month = 	 {23--29 Jul},
  publisher =    {PMLR},
  url = 	 {https://proceedings.mlr.press/v202/gao23f.html}
}
\bibliographystyle{iclr2027_conference}

\clearpage

\appendix
\etocdepthtag.toc{mtappendix}
\etocsettagdepth{mtmain}{none}
\etocsettagdepth{mtappendix}{subsection}
\etocsettocstyle{\begingroup\noindent{\large\bfseries Appendix Contents}\par\vspace{4pt}\small\setlength{\parskip}{0pt}}{\par\vspace{6pt}\noindent\rule{\linewidth}{0.4pt}\par\endgroup\vspace{4pt}}
\etocsetstyle{section}{}{\par\vspace{3pt}\noindent}{\makebox[1.5em][l]{\bfseries\etocnumber}\textbf{\etocname}\hfill\textbf{\etocpage}\par}{}
\etocsetstyle{subsection}{}{\noindent\hspace*{1.5em}}{\makebox[2.3em][l]{\etocnumber}\etocname\leaders\hbox to 0.7em{\hss.\hss}\hfill\etocpage\par}{}
\tableofcontents

\section{Benchmark Details}

\subsection{Games, objectives, and progress potentials}
\label{sec:catalogue-details}
Table~\ref{tab:environment-catalogue} lists the games and objectives.
\begin{table}[ht]
\centering
\small
\setlength{\tabcolsep}{4pt}
\caption{Games and objectives. Each game--difficulty pair is one objective; level counts follow the listed difficulty order. The completion condition applies to each level.}
\label{tab:environment-catalogue}
\begin{tabular}{@{}p{0.19\linewidth}p{0.23\linewidth}rp{0.36\linewidth}@{}}
\toprule
Game & Difficulties & Levels & Level completion \\
\midrule
Bolt Unscrew & Easy / Hard & 8 / 8 & Remove every board. \\
Rush Hour & Easy / Medium / Hard & 10 / 10 / 10 & Move the target vehicle through the exit. \\
Nut and Bolt & Easy / Medium / Hard / Extreme / Nightmare & 3 / 3 / 3 / 3 / 1 & Make every nonempty stack full and monochromatic. \\
Truck Escape & Default & 5 & Remove all vehicles. \\
Maze Paint & Easy / Medium / Hard & 10 / 10 / 10 & Paint every traversable cell. \\
Color Connect & Easy / Hard & 10 / 10 & Connect every matching color pair with non-overlapping paths. \\
\midrule
Total & 16 objectives & 114 & \\
\bottomrule
\end{tabular}
\end{table}

\noindent\textbf{Controls.} Bolt Unscrew uses clicks to select and relocate screws. Rush Hour uses drags to move vehicles along their axes. Nut and Bolt uses a source click and a destination click to transfer the top run of same-colored nuts, subject to capacity and color. Truck Escape uses clicks to request vehicle departures, Maze Paint uses directional swipes, and Color Connect uses clicks to construct paths. Every game provides native restart and menu controls, and Nut and Bolt also provides undo. The agent observes the full browser viewport, including game menus, and the game's own instructions and feedback stay visible throughout.

\noindent\textbf{Progress potentials.} FGP (Appendix~\ref{sec:metrics}) credits partial progress through a game-specific potential $P_g(s)\in[0,1]$ that the evaluator computes privately for each observed state $s$. Bolt Unscrew uses $(b_{\mathrm{exit}}+0.5\,b_{\mathrm{released}})/b_{\mathrm{total}}$, where released boards have no remaining supports but have not exited. Rush Hour combines estimated distance and structural progress with weights 0.8 and 0.2 when a canonical reference path is available, and blocker clearance and target advance with weights 0.7 and 0.3 otherwise; unfinished values are capped at 0.99. Nut and Bolt sums, over colors, the largest homogeneous stack of each color and divides by stack capacity times the number of colors. Truck Escape uses the fraction of vehicles removed, Maze Paint $(n_{\mathrm{painted}}-1)/(n_{\mathrm{paintable}}-1)$, which excludes the initially painted start cell, and Color Connect the fraction of completed pairs. A solved level scores one.
\subsection{Construction, validation, and refresh}
\label{sec:construction-details}
\noindent\textbf{Construction.} A coding model (GPT-5.6-Sol with \texttt{xhigh} reasoning) reproduces executable games and levels from screenshots or other visual references. Part of the suite builds on an existing game implementation, to which we add task data, private evaluation, and fixed launch parameters; Maze Paint and Color Connect are new. Every game implements a browser evaluation bridge with private completion and progress signals and supports deterministic selection of game, difficulty, level, and seed.
\noindent\textbf{Validation.} GUI agents play each candidate build to check controls, state transitions, and game behavior; a defect leads to repair and a new playtest, or to exclusion of the candidate. Solvers and reference replays add checks where implemented, and manual play verifies that every released level is solvable. For Maze Paint, a solver solves all 30 levels, replays each solution to completion, and checks the exact minimum number of slides: 88, 140, and 217 in total for Easy, Medium, and Hard (7--12, 13--16, and 18--25 per level). These counts exclude game entry, transitions, exploration, and recovery.

\noindent\textbf{Refresh.} Validated games and levels are added between evaluation snapshots, and comparisons within a snapshot fix its catalogue, runtime, and seed. Refresh limits repeated exposure to a fixed catalogue but does not by itself establish freedom from contamination: a game reproduced from screenshots can retain familiar mechanics or layouts, and a post-training evaluation would also require the release dates of the added instances and the training cutoff of each model.

\subsection{Benchmark characterization statistics}
\label{sec:characterization}
These definitions underlie Figure~\ref{fig:benchmark-overview}b--c and the corresponding statistics in Section~\ref{sec:environments}.

\noindent\textbf{Human reference.} One human player completed each of the 16 objectives under the evaluation protocol (same start screen, seed, viewport, native GUI actions, and 60-minute budget), without private state, solver output, or external computation. Actions count pointer presses, one per click, drag, or swipe; pointer-move samples are excluded (Table~\ref{tab:human-horizon}). The human issued 99 restarts and 55 undos, on 9 of the 16 objectives.

\begin{table}[ht]
\centering
\small
\caption{Human reference episodes: target levels, pointer presses, and recovery actions per objective.}
\label{tab:human-horizon}
\begin{tabular}{@{}lrrrr@{}}
\toprule
Objective & Levels & Pointer presses & Restarts & Undos \\
\midrule
Bolt Unscrew / Easy & 8 & 100 & 1 & 0 \\
Bolt Unscrew / Hard & 8 & 2,500 & 52 & 0 \\
Rush Hour / Easy & 10 & 89 & 0 & 0 \\
Rush Hour / Medium & 10 & 102 & 0 & 0 \\
Rush Hour / Hard & 10 & 442 & 12 & 0 \\
Nut and Bolt / Easy & 3 & 45 & 0 & 0 \\
Nut and Bolt / Medium & 3 & 169 & 1 & 1 \\
Nut and Bolt / Hard & 3 & 405 & 2 & 3 \\
Nut and Bolt / Extreme & 3 & 538 & 8 & 8 \\
Nut and Bolt / Nightmare & 1 & 1,106 & 19 & 43 \\
Truck Escape / Default & 5 & 134 & 0 & 0 \\
Maze Paint / Easy & 10 & 128 & 0 & 0 \\
Maze Paint / Medium & 10 & 238 & 0 & 0 \\
Maze Paint / Hard & 10 & 476 & 1 & 0 \\
Color Connect / Easy & 10 & 118 & 0 & 0 \\
Color Connect / Hard & 10 & 316 & 3 & 0 \\
\bottomrule
\end{tabular}
\end{table}

\noindent\textbf{Dead-end certification.} In Bolt Unscrew, the evaluator certifies a deadlock (\texttt{no\_available\_hole}) when the board is stable, no operation is pending, and no admissible source--target screw relocation remains; the game does not display this state. In Nut and Bolt, every distinct settled board of a level attempt is classified by exhaustive search over the states reachable under the transfer rule, in which the top run of same-colored nuts moves onto an empty bolt or a matching top nut, as many as fit; every visited board was classified within the search limit. An attempt reaches a dead end if it reaches an unsolvable board. Its delay counts transfers from the first unsolvable board, including the transfer that produced it, until an undo restores a solvable board or the attempt ends. Snapshots record settled boards, so transfers between two snapshots can be missed and delays are lower bounds. Of the 18 human dead-end attempts in Nut and Bolt, 16 ended in a restart and 2 in an undo; of the 83 agent attempts, 51 ended in a restart, 29 in an undo, and 3 at the end of the episode. The other four games have no certificate, and their states are not annotated as dead ends.

\noindent\textbf{Agent statistics.} Agent counts pool the 384 episodes of Table~\ref{tab:main-results} (both settings, all 14 agents), in which agents issued 996 restarts and 384 undos.

\subsection{Criteria for the benchmark comparison}
\label{sec:comparison-criteria}
\noindent\textbf{Scope.} Table~\ref{tab:benchmark-comparison} includes representative benchmarks from four paradigms adjacent to long-horizon visual reasoning: static visual puzzles, GUI task completion, interactive games, and interactive puzzles. Text-only game and puzzle benchmarks discussed in Section~\ref{sec:related-work}, such as SmartPlay and EscapeBench, are omitted because they do not require interpreting a rendered state \citep{ICLR2024_063d1fad,qian-etal-2025-escapebench}. Because its play is self-paced, LongPuzzleBench does not test the latency-sensitive control that real-time play in VideoGameBench and GameWorld also evaluates \citep{zhang2026videogamebenchvisionlanguagemodelscomplete,ouyang2026gameworldstandardizedverifiableevaluation}. Each entry follows the published task definition and default protocol of the cited benchmark; $\dagger$ marks a property present only in a subset of tasks or in one evaluation track, and -- marks a property that does not apply or is not reported.

\noindent\textbf{Property definitions.} \emph{Visual puzzle state}: success is defined by rule constraints on a rendered environment state, and the action sequence must be derived from those constraints rather than read from the task instruction. \emph{Delayed consequences}: actions update a persistent state over many steps, and whether the task succeeds depends on changes made many steps earlier, not only on the next few decisions. \emph{Dead-end states}: the task rules admit legal actions after which the goal is unreachable without undo or restart; irreversible side effects that occur incidentally in application tasks do not qualify. In LongPuzzleBench, this holds in Bolt Unscrew and Nut and Bolt (7 of 16 objectives); in the other four games, moves are reversible or only remove pieces. \emph{Agent-owned recovery}: after a failure, the agent itself decides whether and when to undo, restart, or revise within the episode, without a scripted runtime reset; single-attempt episodes are marked $\times$, and settings without dead ends or attempts are marked --. \emph{Native GUI actions}: the environment changes only through coordinate-based mouse, touch, or key inputs on a rendered application, excluding element-ID commands, abstract game actions, and programmatic state access.

\noindent\textbf{Per-benchmark notes.} PuzzleBench poses questions about a presented configuration, so no action changes the state and the protocol columns do not apply. VisualWebArena and AndroidWorld tasks involve dependent steps on persistent application state, but they are specified by instructions rather than puzzle constraints, and their definitions do not target dependencies that surface many steps later. VisualWebArena acts through element-ID commands on set-of-marks annotations. BALROG provides visual observations in its vision-language track; its puzzle entry reflects BabaIsAI, one of its environments. VideoGameBench and GameWorld contain puzzle games within broader game suites. Some VideoGameBench games use scripted startup, and GameWorld continues after terminal failure through a runtime reset; both therefore take part of the session lifecycle out of the agent's control. GameWorld supports native inputs in a subset of its settings. VLATIM evaluates puzzles designed for a single attempt. ARC-AGI-3 acts through abstract game actions; its agents restart a level after GAME\_OVER through a RESET action, so recovery is agent-owned. AI GameStore is omitted because its generated games with short episodes add no evaluation regime beyond the game benchmarks listed \citep{ying2026aigamestorescalableopenended}.

\noindent\textbf{Instance renewal.} Renewal concerns evaluation maintenance rather than the demands of the task, so Table~\ref{tab:benchmark-comparison} omits it. PuzzleBench, AndroidWorld, and BALROG generate new instances from fixed task families; AI GameStore adds games through synthesis and human refinement; the other benchmarks use fixed task sets, and ARC-AGI-3 reports public and private sets without a refresh protocol. LongPuzzleBench adds validated levels and mechanics between snapshots (Appendix~\ref{sec:construction-details}).

\section{Evaluation Protocol and Metrics}

\subsection{Episode protocol}
\label{sec:full-setup}
\begin{figure}[t]
    \centering
    \includegraphics[width=\linewidth]{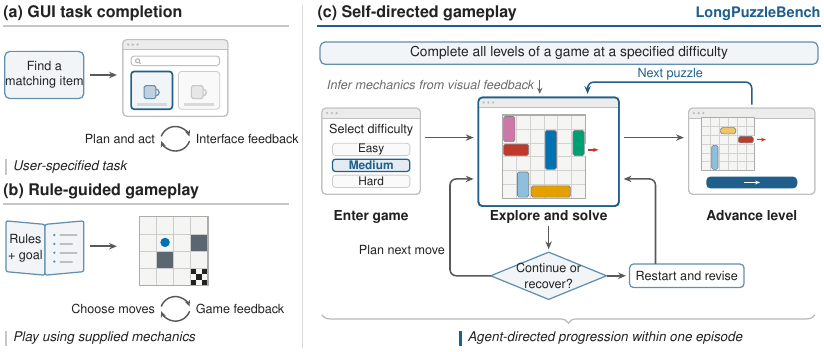}
    \caption{Task guidance and gameplay responsibility in representative settings. (a) An agent uses interface feedback to complete a user-specified application task. (b) Games such as SmartPlay and BALROG supply rules and action descriptions to guide play. (c) LongPuzzleBench supplies a game--difficulty objective; the agent discovers mechanics through visual interaction and controls entry, puzzle solving, recovery, and advancement within one episode.}
    \label{fig:self-directed-gameplay}
\end{figure}

\noindent\textbf{Start and instruction.} Each episode starts at the suite home screen. The agent receives the GUI action specification, the coordinate convention, and the objective instruction of Section~\ref{sec:task-definition}; no manual, solution, navigation procedure, or evaluator signal is supplied. The agent enters the game and decides when to advance, retry, restart, or re-enter (Figure~\ref{fig:self-directed-gameplay}). The player instructions require the agent to act only through the visible interface, to keep playing until the goal is visibly complete, and to end only with an explicit completion declaration.

\noindent\textbf{Budget and termination.} Each objective has a 3,600-second wall-clock budget and no action limit. The clock runs across levels, retries, and re-entry and includes navigation, play, waits, model inference, and transient request retries. An episode ends at full completion, at an explicit final response by the agent, or at the deadline. Success requires evaluator snapshots that show a win on every target level; a completion declaration by the agent is not sufficient. For the GUI-specialized models, the environment answers an unverified completion claim once with a continuation prompt, and a second claim ends the episode.

\noindent\textbf{Scored episodes.} Completions, deadline terminations, and voluntary exits are valid outcomes; a voluntary exit before completion counts as a failure and toward PER. Episodes stopped by provider, browser, environment, logging, or configuration failures, or stopped externally, are not scored. The main results comprise 384 episodes: 160 under Code Execution CUA and 224 with Native GUI Actions (160 from the Codex backbones and 64 from the GUI-specialized models), covering all 16 objectives for every agent in each of its settings.

\subsection{Action settings and agents}
\label{sec:settings-protocol}
Table~\ref{tab:setting-interfaces} summarizes the two settings of Section~\ref{sec:eval-settings}; the human reference plays with native GUI actions under the same protocol (Appendix~\ref{sec:characterization}).

\begin{table}[ht]
\centering\footnotesize
\setlength{\tabcolsep}{5pt}
\caption{The two evaluation settings. Both play the same game in a dedicated Chromium browser at $1280\times900$ and change it only through coordinate-based GUI input (bottom row); they differ in what the agent may do between inputs.}
\label{tab:setting-interfaces}
\begin{tabular}{@{}p{0.47\linewidth}|p{0.47\linewidth}@{}}
\toprule
\multicolumn{1}{c|}{\textbf{Native GUI Actions}} & \multicolumn{1}{c}{\textbf{Code Execution CUA}} \\
\midrule
\textit{Action space: one GUI-action tool} & \textit{Action space: code runtime and GUI input} \\
\texttt{click(x,y)} \hfill primary click & Run JavaScript in a persistent runtime \\
\texttt{double\_click(x,y)} \hfill double click & \quad over screenshots and derived data \\
\texttt{drag(x0,y0,x1,y1)} \hfill drag pointer & Request a screenshot \\
\texttt{press\_key(key)} \hfill game key & Issue the same coordinate GUI inputs, \\
\texttt{type\_text(text)} \hfill text entry & \quad directly or from code \\
\texttt{scroll(direction)} \hfill scroll & Read and write own files \\
\texttt{wait()} \hfill idle & \quad (when file tools are available) \\
\texttt{batch([...])} \hfill ordered actions & \\
\addlinespace[2pt]
Screenshot returned after every call & Screenshots on request \\
Backbones: ten general-purpose (Codex); & Backbones: ten general-purpose (Codex) \\
\quad four GUI-specialized (autonomous harness) & \\
Metrics: OSR, LC, FGP, LPR, H$_{50}$, LC$_\mathrm{hard}$, PER & Metrics: as left, plus NGAR \\
\midrule
\multicolumn{2}{@{}p{0.96\linewidth}@{}}{\textit{Shared in both settings.} Game state changes only through pointer and key input on the rendered page, recorded by the game; one 3,600-second episode clock; no game source, bundled assets, level catalogues, DOM or accessibility data, page variables, hidden state, evaluator outputs, or external retrieval.} \\
\bottomrule
\end{tabular}
\end{table}

\noindent\textbf{Shared constraints.} Both settings use the same objectives, seed-0 layouts, browser and viewport, budget, termination and validity rules, and aggregation. The agent may not read game source or bundled assets, level catalogues, answer files, DOM or accessibility data, page variables, hidden game state, or evaluator outputs, and may not retrieve game information externally. The player instructions state this prohibition in both settings, and every scored trajectory is checked for source or answer access; no episode shows such access.

\noindent\textbf{Native GUI Actions: Codex backbones.} The only action channel is one tool whose calls name the page URL, an action, and its parameters (Table~\ref{tab:setting-interfaces}). Coordinates are integer screenshot pixels. Keys are restricted to letters and game keys (Enter, Tab, Escape, Space, Backspace, and arrows), which excludes browser shortcuts and developer tools, and browser navigation is not exposed. A fixed dispatcher validates each call, applies it to the page, and returns a fresh screenshot; a \texttt{batch} executes its actions in order and returns one screenshot after the last, and each executed action counts separately. The first action must be \texttt{wait}, which returns the initial screenshot and starts the clock. The tool accepts no code, selectors, file paths, or page-state queries. A watchdog checks every call against the schema and the run's URL, and any other tool call or schema violation stops and invalidates the run. The player prompt lists the actions and forbids JavaScript, shell, Python, browser-automation code, DOM manipulation, and any other substitute channel.

\noindent\textbf{Native GUI Actions: GUI-specialized models.} UI-TARS-1.5-7B and the three Qwen3-VL Thinking models run in an autonomous harness with the same action set and validation as the Codex tool, plus a terminal answer. Their serving endpoints do not support this tool, so each model acts in its own output format: Qwen3-VL as text function calls with coordinates on a $0$--$1000$ grid, and UI-TARS-1.5 as \texttt{Thought}/\texttt{Action} text with coordinates on its resized input image. The harness converts both to screenshot pixels, validates them, and dispatches and records one primitive at a time. No code execution, DOM or accessibility data, or browser navigation is available.

\noindent\textbf{Code Execution CUA.} The agent has a persistent JavaScript runtime, exposed through an MCP service, in addition to coordinate GUI input, and may read and write its own files when file tools are available. The instructions permit code execution and leave its use to the agent; they describe initialization, screenshots, and the GUI API without recommending a solving algorithm. Game information must come from returned screenshots. Programs may process these screenshots and any data the agent derives from them, but every change to the game must use the prescribed GUI inputs. Such programs are permitted computation, not privileged access: they lower NGAR but do not invalidate the episode.

\noindent\textbf{Prompts.} Each Codex agent runs as a player with fixed developer instructions for its setting, added to the scaffold's default system prompt, and receives one task message stating the goal, the game and difficulty with their on-screen labels, the page URL, and the setting. The game interface is in Chinese, and so are the Codex prompts; the GUI-specialized models receive the English instruction of Section~\ref{sec:task-definition} with the game's on-screen name, plus a system prompt in their own action format. No prompt contains game rules or strategies. All prompts specify (i) the permitted information: what is visible in the returned screenshots, with the prohibited sources listed above; (ii) the action interface: the tool schema, coordinate convention and bounds, permitted keys, and batch format for Native GUI Actions, and the JavaScript GUI API for Code Execution CUA; (iii) the procedure: select the game and difficulty through the visible interface and check the selection before starting, and use the visible undo or restart controls when needed, treating neither repeated failure nor elapsed time as a reason to stop; and (iv) the ending: once the page shows that the whole objective is complete, send one message that starts with \texttt{BENCHMARK\_COMPLETED} and names the visible evidence. Eighteen episodes (five under Code Execution CUA and 13 with Native GUI Actions, all from GPT-5.6-Luna and the open-weight backbones) used an earlier player prompt whose context also contained text from a since-removed orchestration layer; they pass the same access check.

\begin{table}[ht]
\centering\footnotesize
\setlength{\tabcolsep}{4pt}
\caption{Model access and inference settings. Effort: requested reasoning effort, identical in both settings and matched by the recorded value in every episode. Limit: context window configured in the Codex scaffold (general-purpose) or maximum output tokens per request (GUI-specialized), in thousands of tokens. Price: list price in USD per million uncached input, cached input, and output tokens, retrieved in September 2026 from each model's first-party API or the listed endpoint; DeepSeek 4.1 Flash at peak-hour rates.}
\label{tab:model-settings}
\begin{tabular}{@{}llllr@{}}
\toprule
Agent & Endpoint & Effort & Limit & Price (in / cached / out) \\
\midrule
\multicolumn{5}{@{}l}{\textit{General-purpose, Codex scaffold}} \\
\hspace{0.6em}GPT-6-Astra & OpenAI & medium & 258 & 10.00 / 1.00 / 50.00 \\
\hspace{0.6em}GPT-6-Sol & OpenAI & medium & 258 & 2.00 / 0.20 / 10.00 \\
\hspace{0.6em}GPT-6-Luna & OpenAI & medium & 258 & 0.10 / 0.01 / 0.50 \\
\hspace{0.6em}GPT-5.6-Sol & OpenAI & medium & 258 & 4.00 / 0.40 / 20.00 \\
\hspace{0.6em}GPT-5.6-Terra & OpenAI & medium & 258 & 2.00 / 0.20 / 12.00 \\
\hspace{0.6em}GPT-5.6-Luna & OpenAI & medium & 258 & 0.20 / 0.02 / 1.20 \\
\hspace{0.6em}DeepSeek 4.1 Flash & OpenAI-compatible & high & 996 & 0.30 / 0.006 / 1.20 \\
\hspace{0.6em}Qwen3.8 Flash & OpenAI-compatible & xhigh & 996$^{a}$ / 122$^{b}$ & 0.15 / 0.016 / 0.47 \\
\hspace{0.6em}GLM-5.3-Flash & OpenAI-compatible & max & 950--996 & 0.15 / 0.03 / 0.50 \\
\hspace{0.6em}Kimi-k3 & OpenAI-compatible & max & 996 & 3.00 / 0.30 / 15.00 \\
\addlinespace[2pt]
\multicolumn{5}{@{}l}{\textit{GUI-specialized, autonomous harness}} \\
\hspace{0.6em}UI-TARS-1.5-7B & OpenRouter & -- & 2 & 0.10 / 0.10 / 0.20 \\
\hspace{0.6em}Qwen3-VL-32B-Thinking & SiliconFlow & -- & 8 & 0.20 / -- / 1.50 \\
\hspace{0.6em}Qwen3-VL-30B-A3B-Thinking & SiliconFlow & -- & 8 & 0.29 / -- / 1.00 \\
\hspace{0.6em}Qwen3-VL-8B-Thinking & SiliconFlow & -- & 8 & 0.18 / -- / 2.00 \\
\bottomrule
\multicolumn{5}{@{}l}{\scriptsize $^{a}$Native GUI Actions. $^{b}$Code Execution CUA.}
\end{tabular}
\end{table}

\noindent\textbf{Inference and history.} Table~\ref{tab:model-settings} lists the endpoint and settings of each agent. The Codex scaffold (Codex CLI 0.154--0.156) sets only the reasoning effort; other sampling parameters and output limits are provider defaults. Screenshots are $1280\times900$ PNG images. Codex agents keep the full interaction history until it fills the configured context window; the scaffold then replaces earlier turns with a model-written summary and continues. This compaction occurs in 42 of the 320 Codex episodes, 15 of them Qwen3.8 Flash episodes under Code Execution CUA, whose configured window is the smallest. The GUI-specialized models are queried through the Chat Completions API without streaming and with provider-default sampling; each request replays the full text history with the two most recent screenshots. No auxiliary learned model is used. The game suite and browsers run locally on one workstation (Apple M5 Pro), and all models are queried through remote APIs. The episodes of Table~\ref{tab:main-results} were collected between 16 and 26 September 2026.

\noindent\textbf{Comparability.} Beyond the action channel, the two settings differ in tool-integration latency, and the GUI-specialized models also run in a different harness. Differences between settings, or between general-purpose and GUI-specialized agents, therefore characterize each agent system as a whole (Section~\ref{sec:native-gui-analysis}).

\subsection{Metric definitions}
\label{sec:metrics}
For objective $o$ with target levels $L_o$, let $c_{o\ell}\in\{0,1\}$ indicate an observed win on level $\ell$ in the episode. A won level counts once and keeps its credit across retries; an unvisited level counts as zero. Unless noted otherwise, every metric is computed per objective, averaged equally over the objectives of each game, and then equally over the six games.

\noindent\textbf{Completion.} Objective success and \emph{level completion} are
\begin{equation}
    s_o=\mathbf{1}\!\left[\textstyle\sum_{\ell\in L_o}c_{o\ell}=|L_o|\right],
    \qquad \mathrm{LC}_o=\frac{1}{|L_o|}\sum_{\ell\in L_o}c_{o\ell},
\end{equation}
and OSR averages $100\,s_o$. \emph{Fine-grained progress} also credits partial solutions. Let $b_{o\ell}$ be the maximum potential $P_g$ observed on level $\ell$ (Appendix~\ref{sec:catalogue-details}), set to one for a won level and zero for an unvisited one; then $\mathrm{FGP}_o=|L_o|^{-1}\sum_{\ell\in L_o}b_{o\ell}$. Because levels are won in catalogue order, FGP exceeds LC only through progress on the level where the episode stops. \emph{Hard-tier level completion} (LC$_\mathrm{hard}$) restricts LC to the objectives above each game's easiest tier: Bolt Unscrew Hard, Rush Hour Medium and Hard, Nut and Bolt Medium through Nightmare, Maze Paint Medium and Hard, and Color Connect Hard. Truck Escape has a single tier and is excluded.

\noindent\textbf{Horizon.} Let $v_{e\ell}$ indicate that level $\ell$ was reached in episode $e$, meaning that a ready gameplay state for the requested game, difficulty, and level was observed. Pooling the episodes $E_o$ of objective $o$, the \emph{level pass rate} is
\begin{equation}
    \mathrm{LPR}_o=\frac{\sum_{e\in E_o}\sum_{\ell\in L_o}c_{e\ell}}{\sum_{e\in E_o}\sum_{\ell\in L_o}v_{e\ell}},
\end{equation}
which is undefined, for the objective and overall, when no target level is reached. The \emph{50\% level horizon} uses the eight ten-level objectives $O_{10}$ (Rush Hour, Maze Paint, and Color Connect). With $d_e$ the number of levels solved in episode $e$,
\begin{equation}
    S(k)=\frac{1}{|O_{10}|}\sum_{o\in O_{10}}\frac{1}{|E_o|}\sum_{e\in E_o}\mathbf{1}[d_e\ge k],\qquad
    \mathrm{H}_{50}=\max\{k\in\{0,\dots,10\}:S(k)\ge 0.5\},
\end{equation}
reported as 10+ when $S(10)\ge 0.5$.

\noindent\textbf{Behavior.} The \emph{premature exit rate} (PER) is the percentage of scored episodes that end, without objective success, through a final response or stop action by the agent while budget remains and the interface is usable. Both an incorrect completion declaration and an explicit decision to abandon count; timeouts and evaluator or environment stops do not. The \emph{setting mismatch rate} (SMR) is the percentage of episodes whose evaluator snapshots show gameplay in a game or difficulty other than the requested one before completion or termination; such an episode stays in every denominator, and only progress on the requested levels earns credit. PER and SMR pool episodes within each displayed scope.

\noindent\textbf{Native GUI action ratio.} NGAR is $100\,N_{\mathrm{native}}/(N_{\mathrm{native}}+N_{\mathrm{other}})$, pooled over all model action steps of the Code Execution CUA episodes in scope; a step is one tool call or an explicit terminal answer. A step is native when it only observes the screen or issues coordinate-based mouse, touch, or key input, including fixed coordinate batches; a call counts once even if it contains several inputs or an input fails. Offline solving, programmatic image analysis, helper or module authoring, file and coordination tools, standalone waits, terminal answers, semantic element targeting, synthetic DOM events, and direct DOM, accessibility, canvas, or game-state access are non-native, as is any call that combines computation with GUI input. Internal reasoning, commentary, tool results, and automatic recorder events are not steps. NGAR is not reported with Native GUI Actions, where every accepted action is native by construction.

\subsection{Compute and cost}
\label{sec:compute-cost}
Table~\ref{tab:usage-cost} reports the tokens consumed by the 384 episodes of Table~\ref{tab:main-results} and their cost at the list prices of Table~\ref{tab:model-settings}. The episodes issue 45{,}997 model requests with 5.65 billion input tokens, 97.7\% of them cache hits, and 30.7 million output tokens, for an estimated USD 868 (571 with Native GUI Actions and 297 under Code Execution CUA). Input dominates because every request resends the interaction history, including screenshots, and prompt caching bills most of it at the cached rate. The controlled diagnostics of Section~\ref{sec:analysis} are not included.

\begin{table}[ht]
\centering\footnotesize
\setlength{\tabcolsep}{5pt}
\caption{Token usage and estimated API cost of the 384 episodes of Table~\ref{tab:main-results}. Input counts all prompt tokens, including cache hits; output includes reasoning tokens. Cost applies the list prices of Table~\ref{tab:model-settings} to uncached input, cached input, and output; it is an estimate, not a billed amount.}
\label{tab:usage-cost}
\begin{tabular}{@{}lrrrrr@{}}
\toprule
Agent & Requests & Input (M) & Cached (\%) & Output (M) & Cost (USD) \\
\midrule
\multicolumn{6}{@{}l}{\textit{Native GUI Actions}} \\
\hspace{0.6em}GPT-6-Astra & 1{,}286 & 100.7 & 97.9 & 0.19 & 129.70 \\
\hspace{0.6em}GPT-6-Sol & 2{,}592 & 237.5 & 98.6 & 0.48 & 58.41 \\
\hspace{0.6em}GPT-6-Luna & 2{,}713 & 288.4 & 98.2 & 0.37 & 3.53 \\
\hspace{0.6em}GPT-5.6-Sol & 3{,}077 & 327.5 & 98.3 & 0.55 & 161.87 \\
\hspace{0.6em}GPT-5.6-Terra & 2{,}132 & 218.6 & 98.1 & 0.44 & 56.32 \\
\hspace{0.6em}GPT-5.6-Luna & 2{,}411 & 233.2 & 97.8 & 0.32 & 5.97 \\
\hspace{0.6em}DeepSeek 4.1 Flash & 3{,}431 & 1{,}125.9 & 99.5 & 5.33 & 14.85 \\
\hspace{0.6em}Qwen3.8 Flash & 1{,}503 & 233.1 & 97.4 & 2.16 & 5.56 \\
\hspace{0.6em}GLM-5.3-Flash & 2{,}519 & 508.8 & 98.1 & 1.01 & 16.94 \\
\hspace{0.6em}Kimi-k3 & 1{,}415 & 213.7 & 97.8 & 1.32 & 96.91 \\
\hspace{0.6em}UI-TARS-1.5-7B & 706 & 5.0 & 53.9 & 0.09 & 0.51 \\
\hspace{0.6em}Qwen3-VL-32B-Thinking & 1{,}000 & 6.0 & 0.0 & 1.57 & 3.55 \\
\hspace{0.6em}Qwen3-VL-30B-A3B-Thinking & 2{,}184 & 18.1 & 0.0 & 3.13 & 8.38 \\
\hspace{0.6em}Qwen3-VL-8B-Thinking & 1{,}465 & 10.1 & 0.0 & 3.37 & 8.56 \\
\hspace{0.6em}\textit{Subtotal} & 28{,}434 & 3{,}526.5 & 97.5 & 20.32 & 571.07 \\
\addlinespace[2pt]
\multicolumn{6}{@{}l}{\textit{Code Execution CUA}} \\
\hspace{0.6em}GPT-6-Astra & 1{,}131 & 54.8 & 97.2 & 0.10 & 73.58 \\
\hspace{0.6em}GPT-6-Sol & 1{,}213 & 68.7 & 97.4 & 0.22 & 19.16 \\
\hspace{0.6em}GPT-6-Luna & 2{,}501 & 261.3 & 98.2 & 0.55 & 3.30 \\
\hspace{0.6em}GPT-5.6-Sol & 1{,}468 & 116.8 & 97.9 & 0.40 & 63.51 \\
\hspace{0.6em}GPT-5.6-Terra & 1{,}612 & 134.1 & 97.6 & 0.40 & 37.41 \\
\hspace{0.6em}GPT-5.6-Luna & 1{,}217 & 86.3 & 96.4 & 0.24 & 2.57 \\
\hspace{0.6em}DeepSeek 4.1 Flash & 2{,}624 & 714.1 & 99.4 & 4.04 & 10.32 \\
\hspace{0.6em}Qwen3.8 Flash & 1{,}653 & 97.6 & 92.0 & 2.19 & 3.64 \\
\hspace{0.6em}GLM-5.3-Flash & 2{,}617 & 445.3 & 98.2 & 1.11 & 14.90 \\
\hspace{0.6em}Kimi-k3 & 1{,}527 & 142.9 & 97.7 & 1.13 & 68.74 \\
\hspace{0.6em}\textit{Subtotal} & 17{,}563 & 2{,}121.9 & 98.1 & 10.38 & 297.13 \\
\addlinespace[2pt]
\midrule
Total & 45{,}997 & 5{,}648.4 & 97.7 & 30.70 & 868.19 \\
\bottomrule
\end{tabular}
\end{table}

\section{Extended Main Results}

\subsection{Per-objective results}
\label{sec:settings-results}
\begin{figure}[ht]
\centering
\includegraphics[width=\linewidth]{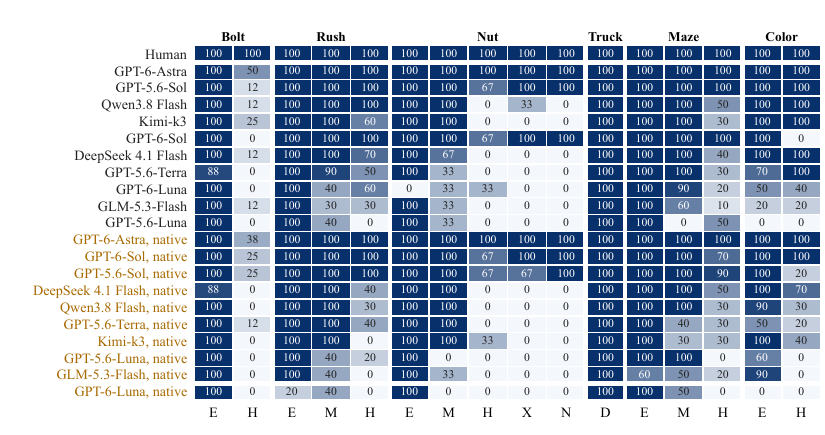}
\caption{Level completion (\%) per objective for the human reference and the ten general-purpose backbones under Code Execution CUA (black labels) and with Native GUI Actions (orange labels), each group ordered by LC$_\mathrm{hard}$. E/M/H: Easy/Medium/Hard; X: Extreme; N: Nightmare; D: Default.}
\label{fig:objective-heatmap}
\end{figure}

Figure~\ref{fig:objective-heatmap} gives level completion per objective for every general-purpose agent in both settings; Appendix~\ref{sec:expanded-results} lists every episode's outcome and the denominators behind Table~\ref{tab:main-results}. Difficulty tiers are comparable within a game but not across games, whose layouts differ. The OSR losses of DeepSeek 4.1 Flash and Kimi-k3 without code (Section~\ref{sec:native-gui-analysis}) fall on different objectives. Kimi-k3 solves Maze Paint Medium and Color Connect Hard only under Code Execution CUA (LC 100\% against 30\% and 40\% with Native GUI Actions), and its Rush Hour Hard LC falls from 60\% to 0\% without code. DeepSeek 4.1 Flash solves Bolt Unscrew Easy and Color Connect Hard only with code (100\% against 87.5\% and 70\%) but Nut and Bolt Medium only without it (100\% against 66.7\%).

\subsection{Behavioral sources of the gap between settings}
\label{sec:setting-mechanism}
This analysis pairs the Native GUI Actions and Code Execution CUA episodes of the ten Codex backbones by backbone and objective (160 pairs). It uses every tool call with its arguments and timestamps, the pointer and key inputs and puzzle states that the game records in both settings, and per-level win times from evaluator snapshots; reasoning content is encrypted and is not analyzed. A Code Execution CUA call is a \emph{computation call} when its code does more than list GUI inputs, screenshot requests, and waits: it searches for moves, analyzes data, or generates inputs through loops, computed coordinates, or helpers defined in earlier calls. This classification differs from NGAR, which also counts standalone waits and terminal answers as non-native. \emph{Matched levels} are the second and later levels that one backbone solves in both settings; levels are solved in order, so matching by level is exact. Table~\ref{tab:setting-mechanism} summarizes the measurements.

\begin{table}[ht]
\centering\small
\setlength{\tabcolsep}{4pt}
\caption{Paired behavioral comparison of the two settings over the ten Codex backbones (160 episodes per setting). Inputs are pointer and key inputs recorded by the game. Matched-level ratios are medians of per-level ratios of Code Execution CUA to Native GUI Actions for time, calls, screenshots, and executed inputs. Drag slips are non-adjacent jumps and wrong-endpoint releases.}
\label{tab:setting-mechanism}
\begin{tabular}{@{}lcc@{}}
\toprule
& Native GUI Actions & Code Execution CUA \\
\midrule
\multicolumn{3}{@{}l}{\textit{Outcomes}} \\
Episodes solved / timed out / abandoned & 84 / 41 / 35 & 92 / 31 / 37 \\
Timeouts that solve a level in the final 15 min & 15 of 41 & 7 of 31 \\
Level restarts per episode (mean) & 2.4 & 3.6 \\
\addlinespace[2pt]
\multicolumn{3}{@{}l}{\textit{Interaction}} \\
Tool calls per episode (median) & 100.0 & 80.5 \\
Executed inputs per call & 1.91 & 2.66 \\
Screenshots returned per call & 1.00 & 0.81 \\
Observation-only calls (\%) & 18.0 & 11.9 \\
Computation calls (\% of calls / \% of inputs) & -- & 20.1 / 56.3 \\
Solved levels with ${\geq}80\%$ of inputs from one call (\%) & 4.9 & 17.4 \\
Model / tool time per call, median (s) & 6.2 / 0.30 & 6.9 / 0.63 \\
Levels solved per hour & 9.4 & 12.0 \\
\addlinespace[2pt]
\multicolumn{3}{@{}l}{\textit{Matched levels: Code Execution CUA / Native GUI Actions ratio (time / calls / screenshots / inputs)}} \\
All 591 matched levels & \multicolumn{2}{c}{0.81 / 0.83 / 0.67 / 0.93} \\
279 levels with computation calls & \multicolumn{2}{c}{0.60 / 0.74 / 0.50 / 0.92} \\
312 levels with GUI calls only & \multicolumn{2}{c}{0.94 / 1.00 / 0.80 / 1.00} \\
\addlinespace[2pt]
\multicolumn{3}{@{}l}{\textit{Operation errors recorded by the game (\%)}} \\
Maze Paint: invalid / zero-progress / repeated swipes & 17.7 / 25.9 / 4.1 & 12.9 / 29.8 / 9.7 \\
Color Connect: invalid gestures / drag slips & 5.3 / 5.3 & 6.1 / 4.4 \\
Tool outputs flagged as errors & 0.1 & 1.3 \\
\bottomrule
\end{tabular}
\end{table}

\noindent\textbf{Where the gap arises.} Pooled over the ten backbones, Code Execution CUA solves 92 of 160 objectives and Native GUI Actions 84. Pair by pair, Code Execution CUA solves more levels in 38 pairs and fewer in 17. Its gains come from episodes that end unsolved with Native GUI Actions: of the 41 timeouts, it solves more levels in 20 and completes 11; of the 35 voluntary exits, it solves more levels in 18 and completes 4. Its losses come mainly from its own exits: in 6 of the 7 pairs where only the Native GUI Actions episode succeeds, the Code Execution CUA agent stops voluntarily.

\noindent\textbf{Efficiency comes from computation, not from the interface.} On the 591 matched levels, Code Execution CUA takes a median 68.0~s per level against 94.0~s, with 7 against 8 calls and 6 against 8 screenshots, but a similar number of executed inputs (14 against 17). On the 279 levels where the Code Execution CUA agent computes, the median ratio is 0.60 for time, 0.74 for calls, and 0.50 for screenshots; on the 312 levels with GUI calls only, calls and inputs are unchanged (1.00) and time falls only slightly (0.94), so the JavaScript interface alone brings little speedup. With computation, time per level falls for all nine backbones that have such levels (median ratios 0.13--0.79).

\noindent\textbf{One call carries a computed sub-plan.} Computation calls are 20.1\% of Code Execution CUA calls but issue 56.3\% of its executed inputs; a call that generates inputs by program issues 13.3 on average, against 5.8 for a batch with Native GUI Actions. One call issues at least 80\% of a solved level's inputs in 17.4\% of Code Execution CUA levels, against 4.9\% with Native GUI Actions (30.4\% against 18.0\% at a 50\% threshold). With Native GUI Actions, 22.5\% of calls batch literal coordinates, but 59.5\% are single inputs and 18.0\% only observe. Code Execution CUA agents turn a transcribed board into inputs in three recurring ways: a breadth- or depth-first search over the board (in Maze Paint, Nut and Bolt, and Color Connect), a replay loop that converts a move list to pixels through a cell-to-coordinate map, and move lists or helpers kept in the persistent runtime across calls. On Maze Paint Hard, for example, GPT-5.6-Sol with Native GUI Actions spends 47--84 calls per level, half of them observation only, and reaches the deadline on level~10 with nine solved; under Code Execution CUA, most levels after the first take three or four calls (transcribe and search, replay the 18--44 swipes, advance), and it solves all ten in 1,068~s with 56 calls.

\noindent\textbf{Efficiency becomes completion under the deadline.} Faster levels change the outcome only when the Native GUI Actions episode runs out of time while still solving. In 10 of the 11 such timeouts that Code Execution CUA completes, the agent with Native GUI Actions solved its last level within 13.2 minutes of the deadline (29.4 minutes in the remaining episode) and had solved 60\% of the target levels on average; the Code Execution CUA episodes finish in a median 1,093~s (666--2,685~s) with a median 94 calls against 268, and 9 of the 11 spend at least 20\% of their calls on computation. Across backbones, Kimi-k3 and Qwen3.8 Flash each time out on eight objectives with Native GUI Actions and compute in 15.6\% and 40.0\% of their Code Execution CUA calls; all 20 and 19 levels they add come from those objectives. GLM-5.3-Flash times out on ten objectives but computes in only 9.2\% of calls; it adds two levels overall but loses one on those objectives. GPT-6-Astra and GPT-6-Sol compute often (34.7\% and 38.7\% of calls) but rarely time out with Native GUI Actions; Code Execution CUA raises their pace from 15.1 to 24.9 and from 14.9 to 27.4 levels per hour without raising completion.

\noindent\textbf{Exits and restarts.} With Native GUI Actions, failures divide into 41 timeouts, 32 of them from the four open-weight backbones, and 35 voluntary exits. Under Code Execution CUA, timeouts fall to 31 while exits rise to 37, 33 of them from GPT-6-Luna, GPT-5.6-Luna, and GPT-5.6-Terra, which compute in 3.3--15.8\% of their calls; GPT-5.6-Luna exits in 11 of 16 episodes and solves 12 fewer levels than with Native GUI Actions. Exit timing also accounts for the 28 levels that GPT-6-Luna adds under Code Execution CUA while computing in 3.3\% of its calls: its Native GUI Actions episodes stop earlier (on Color Connect Easy, after 30 calls with no level solved, against 171 calls and five levels). Cheap replay also makes restarts cheap, and restarts rise from 2.4 to 3.6 per episode; on Nut and Bolt Medium, which DeepSeek 4.1 Flash completes in 1,585~s with Native GUI Actions, its Code Execution CUA agent restarts 22 times and times out on the third level.

\noindent\textbf{Alternative explanations.} Tool latency does not explain the gain: tool time per call is higher under Code Execution CUA (median 0.63~s against 0.30~s), as is model time per call (6.9~s against 6.2~s). Operation errors do not favor either setting consistently. In Color Connect, invalid gestures change little (5.3\% and 6.1\%) and drag slips fall from 5.3\% to 4.4\%; in Maze Paint, invalid swipes fall from 17.7\% to 12.9\%, but zero-progress swipes rise from 25.9\% to 29.8\% and repeated swipes from 4.1\% to 9.7\%. Qwen3.8 Flash, which adds 19 levels, already makes few invalid operations with Native GUI Actions (1.4--3.8\% in both games), and tool errors are rare (1.3\% of calls). Kimi-k3 is the exception: its invalid Maze Paint swipes fall from 20.7\% to 2.5\%, so cleaner execution may contribute to its gain in that game.

\noindent\textbf{Interpretation and limits.} Code Execution CUA changes the unit of decision. With Native GUI Actions, each move costs a model turn and usually a screenshot; an agent that has transcribed a board can instead search it and execute a level's solution within one turn. Under a budget shared by all levels, fewer turns per level convert episodes that are still progressing into completions, without stronger visual reasoning: the model must still read the board, and the search runs outside its reasoning (Section~\ref{sec:native-gui-analysis}). Screenshots requested only when needed and replay through coordinate maps contribute secondarily. For Kimi-k3 and Qwen3.8 Flash, model time per call on matched levels is also lower under Code Execution CUA (18.2~s against 25.9~s, and 20.6~s against 23.3~s), and encrypted reasoning does not allow separating shorter deliberation from provider latency. Because the settings also differ in tool integration, these are associations within the evaluated episodes, not controlled estimates of the effect of the action channel.

\subsection{Human--agent trajectory comparison}
\label{sec:human-agent-protocol}
\begin{figure}[ht]
\centering
\includegraphics[width=\linewidth]{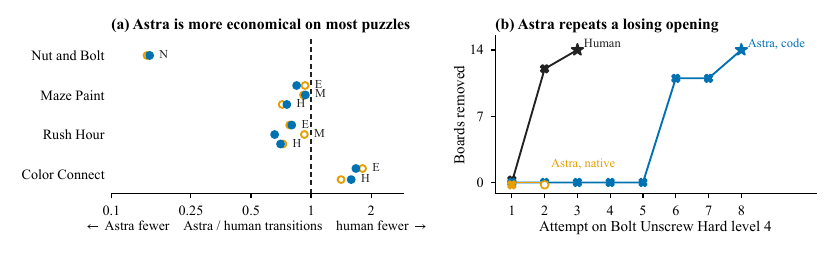}
\caption{Human and GPT-6-Astra succeed on different puzzles (Astra episodes of Table~\ref{tab:main-results}; hollow orange: Native GUI Actions, filled blue: Code Execution CUA). (a) Ratio of Astra's to the human's state transitions per matched objective (E/M/H: Easy/Medium/Hard; N: Nightmare). (b) Boards removed in each attempt on Bolt Unscrew Hard level~4 ($\times$: dead end; star: level cleared); Astra with Native GUI Actions reaches this level two minutes before its deadline. One human trajectory per objective.}
\label{fig:human-astra}
\end{figure}

Section~\ref{sec:human-agent-diagnostic} compares the human reference with the GPT-6-Astra episodes of Table~\ref{tab:main-results} in both settings; with Native GUI Actions, the two share objective, seed, viewport, target levels, native action space, and deadline. Because the human and agent recorders differ, the comparison uses puzzle-state variables that both record, not interface-specific pointer counts, and reports a metric only when both recorders expose its field. In Bolt Unscrew Hard, it follows every attempt on visible level~4, the first level that Astra with Native GUI Actions does not clear, and records boards removed, free holes, and whether the attempt deadlocks or clears the level. Attempts split where the board returns to its initial layout, and snapshots stay in event order because a deadlocked and a restarted state can share a recorder step. Nut and Bolt keeps the last settled snapshot per recorder step and counts changes of the canonical stack configuration, revisits, solved-stack regressions, and resets. Maze Paint and Rush Hour sum the final cumulative moves of each completed level, and Color Connect counts gestures. Screenshots, waits, pointer motions, and unchanged snapshots do not count. Figure~\ref{fig:human-astra} and Table~\ref{tab:human-astra-mechanism-details} report the results.

\begin{table}[!h]
\centering
\footnotesize
\setlength{\tabcolsep}{3pt}
\caption{Trajectory diagnostics for the human reference and the GPT-6-Astra episodes of Table~\ref{tab:main-results} in both settings (Native: Native GUI Actions; Code: Code Execution CUA). Bolt Unscrew values describe visible level~4 (14 boards, four free holes at the start), the first level Astra does not clear with Native GUI Actions, which it reaches near the deadline. Maze Paint and Rush Hour values sum moves over ten levels per objective; E/M/H: Easy/Medium/Hard.}
\label{tab:human-astra-mechanism-details}
\begin{tabular}{@{}llrrr@{}}
\toprule
Puzzle & Diagnostic & Human & Astra, Native & Astra, Code \\
\midrule
Bolt Unscrew Hard L4 & Dead ends with no board removed & 1 & 1 & 5 \\
 & Attempt that clears the level & 3 & -- & 8 \\
 & Most boards in one attempt & 14 & 0 & 14 \\
Nut and Bolt Nightmare & Stack-state changes & 431 & 65 & 67 \\
 & Revisited states & 115 & 4 & 0 \\
 & Solved-stack regressions & 7 & 0 & 0 \\
 & Resets & 17 & 0 & 0 \\
Maze Paint & Moves, E / M / H & 104 / 204 / 381 & 97 / 187 / 273 & 88 / 191 / 288 \\
Rush Hour & Moves, E / M / H & 55 / 67 / 155 & 43 / 62 / 112 & 44 / 44 / 109 \\
Color Connect & Gestures, E / H & 105 / 223 & 190 / 315 & 176 / 354 \\
\bottomrule
\end{tabular}
\end{table}

\section{Diagnostic Studies}

\subsection{Common design and validity rules}
\label{sec:diagnostic-common}
\noindent\textbf{Agent and harness.} The controlled diagnostics use a screenshot-to-action harness with the GUI action set of Native GUI Actions, which differs from the Codex scaffold of Table~\ref{tab:main-results}. At takeover, the agent receives the public action prefix as text, up to four preceding screenshot--action pairs, and the current screenshot; afterwards, it keeps up to 12 recent action turns and four images, with a summary every eight turns. The viewport is $1280\times900$, and the runner waits two seconds after each dispatched action. The studies evaluate GPT-5.6-Sol and repeat key contrasts with GPT-5.6-Terra and GPT-5.6-Luna (Sol, Terra, and Luna in Section~\ref{sec:analysis} and below); the dashed series of Figure~\ref{fig:section5-supports} are these three-model follow-ups.

\noindent\textbf{Matched conditions.} Within a matched block, conditions share the restored state, objective, screenshot, public trajectory prefix, action interface, and continuation limit; only the intervention differs. Checkpoints and source failures were fixed before any continuation outcome was inspected.

\noindent\textbf{Validity.} Agent stops, timeouts, ineffective actions, repeated failures, and noncompletion are valid outcomes. API transport failures, browser failures, corrupted restoration, and incomplete model responses are infrastructure failures; such an attempt is replaced under the same assignment. When a cell has several valid attempts, the first in a fixed order is the primary observation and the others serve only to test stability. Counts in Section~\ref{sec:analysis} include valid runs only.

\noindent\textbf{Reporting.} Analyses report paired counts, state- or source-level effects, ranges, and standard deviations, and do not treat repeated actions within one trajectory as independent samples. Original and follow-up panels are reported separately when they differ in model or unit coverage, claims are scoped to the tested states and games, and null or reversed paired effects are reported as observed.

\subsection{RQ1: progression and move-level measures}
\label{sec:rq1-observed}
The observational measures use the Native GUI Actions episodes of the ten general-purpose backbones in Table~\ref{tab:main-results}, 16 per backbone.

\begin{figure}[ht]
\centering
\includegraphics[width=\linewidth]{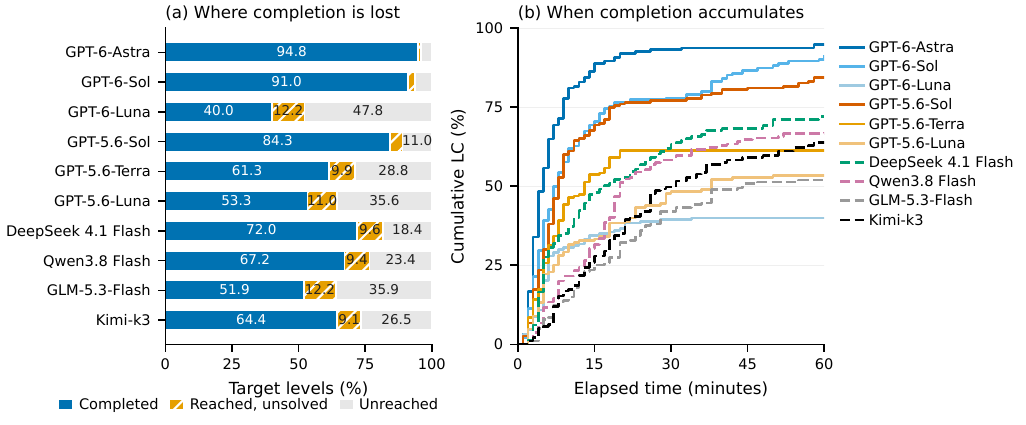}
\caption{Progression of the ten general-purpose backbones with Native GUI Actions (episodes of Table~\ref{tab:main-results}). (a) Each target level is completed, reached but unsolved, or unreached at episode end. Bars average objectives within games and games equally. (b) Cumulative LC uses first-win timestamps and the same weights. Terminated episodes retain their final completion through 60 minutes.}
\label{fig:rq1-progression}
\end{figure}

\noindent\textbf{Reached and unreached levels.} A target level is reached when the evaluator observes its ready gameplay state; each episode contributes one observation per target level. With $r_{o\ell}$ indicating reach and $R_o=|L_o|^{-1}\sum_{\ell}r_{o\ell}$, the LC deficit decomposes as
\[
1-\mathrm{LC}_o=(1-R_o)+(R_o-\mathrm{LC}_o),
\]
with the same game-macro weights for every term. The unreached share is the aggregated $1-R_o$ divided by the aggregated $1-\mathrm{LC}_o$. Across the ten backbones, unreached levels account for 63.7--80.0\% of the deficit (Figure~\ref{fig:rq1-progression}a), and completion conditional on reaching a level ranges from 75.5\% (GPT-6-Luna, 37/49) to 99.1\% (GPT-6-Astra, 109/110; Table~\ref{tab:rq1-denominators}). Because levels are won in order, stalling on one level also forfeits every later one, so a large coverage deficit can originate in a few unresolved puzzles.

\begin{table}[ht]
\centering\small
\setlength{\tabcolsep}{4pt}
\caption{RQ1 denominators and stopping times with Native GUI Actions (Table~\ref{tab:main-results} episodes, 16 per agent). Reached and won levels are pooled episode--level observations. Conditional completion is wins/reached and differs from the macro-weighted LC in Figure~\ref{fig:rq1-progression}. Unreached share: fraction of the LC deficit in levels never reached. Exits/failures counts premature exits among unsuccessful episodes. Exit time: minutes, median [first quartile, third quartile], conditional on premature exit.}
\label{tab:rq1-denominators}
\begin{tabular}{@{}lrrrrr@{}}
\toprule
Agent & Wins/reached & Conditional (\%) & Unreached (\%) & Exits/failures & Exit time \\
\midrule
GPT-6-Astra & 109/110 & 99.1 & 80.0 & 0/1 & N/A \\
GPT-6-Sol & 104/107 & 97.2 & 70.0 & 0/3 & N/A \\
GPT-6-Luna & 37/49 & 75.5 & 79.7 & 11/12 & 19.5 [5.4, 33.7] \\
GPT-5.6-Sol & 97/102 & 95.1 & 70.4 & 1/5 & 20.6 \\
GPT-5.6-Terra & 68/77 & 88.3 & 74.3 & 9/9 & 17.2 [13.7, 21.0] \\
GPT-5.6-Luna & 58/68 & 85.3 & 76.3 & 10/10 & 7.0 [5.5, 8.6] \\
DeepSeek 4.1 Flash & 84/92 & 91.3 & 65.8 & 2/8 & 35.2 [27.1, 43.4] \\
Qwen3.8 Flash & 77/85 & 90.6 & 71.4 & 0/8 & N/A \\
GLM-5.3-Flash & 53/65 & 81.5 & 74.7 & 2/12 & 27.3 [18.8, 35.9] \\
Kimi-k3 & 70/78 & 89.7 & 74.4 & 0/8 & N/A \\
\bottomrule
\end{tabular}
\end{table}

\noindent\textbf{Stopping.} GPT-6-Astra and GPT-6-Sol never stop early; their four unsuccessful episodes reach the deadline. GPT-5.6-Terra and GPT-5.6-Luna end every unsuccessful episode voluntarily (9 of 9 and 10 of 10), at a median of 17.2 and 7.0 minutes, and GPT-6-Luna ends 11 of its 12 at a median of 19.5 minutes. Stopping forfeits later gains: GPT-5.6-Terra leads Qwen3.8 Flash after 15 minutes (53.6\% against 31.5\% LC), but Qwen3.8 Flash keeps playing and finishes higher (67.2\% against 61.3\%; Figure~\ref{fig:rq1-progression}b).

\noindent\textbf{Maze Paint counters.} Maze Paint counters are cumulative within an attempt; level changes and counter resets delimit attempts, and each attempt contributes its terminal record, at the first win or at the last state before a reset or the episode end. Productive moves paint at least one new cell and, with zero-progress moves, partition valid moves; invalid moves form a separate category. A state is the ball position together with the painted-cell configuration, and the revisit denominator pools unique and revisited states, including each attempt's initial state. The 30 episodes contribute 260 attempts, all with complete counters (Table~\ref{tab:rq1-maze}). GPT-6-Astra and GPT-6-Sol revisit an earlier state in 1.0\% and 1.3\% of observations; the three agents that revisit most (33.9--52.4\%) complete 50.0--66.7\% of Maze Paint levels. GLM-5.3-Flash revisits rarely (7.1\%) yet completes 43.3\%, because its slower play reaches the deadline. For Figure~\ref{fig:section5-solvability}c, counter increments between consecutive snapshots are assigned to the coverage bin (fifths of paintable cells) of the earlier snapshot and pooled within bin and outcome; unsolved attempts contribute 460 valid moves and solved attempts 3{,}980.

\begin{table}[t]
\centering\small
\setlength{\tabcolsep}{4pt}
\caption{Maze Paint move diagnostics with Native GUI Actions (three episodes per agent, 260 attempts in total). Productive moves paint at least one new cell. State observations comprise the initial state and the states after game moves, pooled within attempts; revisits repeat an already observed position and painted-cell configuration. Counters include retries and stop at each level win or attempt end.}
\label{tab:rq1-maze}
\begin{tabular}{@{}lrrrr@{}}
\toprule
Agent & Maze LC (\%) & Valid/attempted & Productive/valid (\%) & Revisited/observed (\%) \\
\midrule
GPT-6-Astra & 100.0 & 557/557 & 441/557 (79.2) & 6/587 (1.0) \\
GPT-6-Sol & 90.0 & 505/508 & 386/505 (76.4) & 7/537 (1.3) \\
GPT-6-Luna & 50.0 & 276/385 & 169/276 (61.2) & 143/408 (35.0) \\
GPT-5.6-Sol & 96.7 & 687/714 & 446/687 (64.9) & 64/745 (8.6) \\
GPT-5.6-Terra & 56.7 & 361/541 & 245/361 (67.9) & 192/566 (33.9) \\
GPT-5.6-Luna & 66.7 & 451/863 & 256/451 (56.8) & 466/890 (52.4) \\
DeepSeek 4.1 Flash & 83.3 & 700/833 & 422/700 (60.3) & 203/868 (23.4) \\
Qwen3.8 Flash & 76.7 & 381/396 & 299/381 (78.5) & 25/420 (6.0) \\
GLM-5.3-Flash & 43.3 & 266/277 & 187/266 (70.3) & 21/295 (7.1) \\
Kimi-k3 & 53.3 & 256/323 & 192/256 (75.0) & 69/341 (20.2) \\
\bottomrule
\end{tabular}
\end{table}

\noindent\textbf{Bolt Unscrew capacity.} Figure~\ref{fig:section5-solvability}a--b uses the RQ3 retries, which start from a verified restart of a level's solvable initial board. For each model decision, we take the last stable evaluator state before it: free holes, boards exited, and the deadlock certificate. Consecutive distinct states define moves. A move \emph{frees a board} if boards exited increases, \emph{fills a hole} if free holes decrease without a board exiting, and \emph{deadlocks} if the next state carries the certificate. The reference path is the certified 16-move solution of Hard level~1 that also supplies the RQ2 solvable controls. Of 73 moves made with one free hole, 55 deadlock the board, as do 15 of 17 first moves from one-hole level starts. Stated reasons are the \texttt{Thought} field of each model response, coded with fixed regular expressions; a fatal move \emph{predicts a release} if its rationale mentions releasing, freeing, dropping, or clearing a piece (28 of 55).

\noindent\textbf{Same-state action intervention.} For action $a$ in state $s$, the extra cost is $e(s,a)=1+d(T(s,a))-d(s)$, where $d$ is the exact remaining path length and $T$ the deterministic transition; a shortest-path move has $e=0$. Three Maze Paint states are restored, and from each, one branch executes a previously observed legal action with positive extra cost and the other a legal shortest-path action. Both successors are verified to be solvable, so the intervention tests how one costly move changes the continuation under a finite budget, not whether it destroys solvability. Branch order is balanced, and each state is tested three times with each of Sol, Terra, and Luna, giving 27 paired units (54 continuations) compared on same-level completion.

\subsection{RQ2: certified dead-end takeover}
\label{sec:rq2-diagnostics}
\noindent\textbf{Design.} The original study evaluates GPT-5.6-Sol (medium reasoning effort, temperature~0) on Bolt Unscrew Hard at seed~0. It contains 24 certified dead-end checkpoints (D) from seven source levels (Table~\ref{tab:rq2-source-effects}), each under conditions A, B, and C, and 12 solvable checkpoints (N), distinct states along one verified successful trajectory on level~1, each under A and B: 96 continuations, one per checkpoint and condition. A gives the ordinary objective and visual context, B adds truthful state information, and C adds the label of the restart button to B on D (Table~\ref{tab:rq2-prompts}). The agent chooses every action; the controller only prepares the checkpoint. A-to-B thus tests an explicit state diagnosis and B-to-C the added naming of the recovery control. D and N are separate cohorts, so differences between them describe the selected states rather than a paired intervention.

\noindent\textbf{Checkpoints.} A D checkpoint is an unfinished, stable board carrying the \texttt{no\_available\_hole} certificate; independent replays verify the board and screenshot, and a native restart must restore a playable board on the same level. An N checkpoint has a verified normal-play completion that needs fewer GUI actions than restarting and replaying the solution, and passes two independent legal replays to same-level completion. Certificates and solutions stay private. The controller replays the legal prefix with fixed simulation steps and holds the game until the initial screenshot is captured, so paired continuations start from identical boards and pixels. The intervention text enters the instruction at takeover and stays in context, including after a restart. Conditions follow a fixed interleaved schedule within checkpoint blocks.

\begin{table}[ht]
\centering
\small
\caption{RQ2 intervention text. A appends no message. The restart button carries a Chinese label, given here by its English translation.}
\label{tab:rq2-prompts}
\begin{tabular}{@{}p{0.15\linewidth}p{0.80\linewidth}@{}}
\toprule
Condition/state & Appended information \\
\midrule
B/D & State information: The current puzzle state cannot be completed through normal forward play. Recovery is required before progress can continue. \\
B/N & State information: The current puzzle state remains solvable through normal forward play. \\
C/D & B's D message, followed by: Recovery information: Use the visible button labelled Restart to restart the current level. \\
\bottomrule
\end{tabular}
\end{table}

\noindent\textbf{Endpoints.} The takeover clock includes inference, waits, and GUI operations. A \emph{timely restart} requires a recovery action within the first ten actions and 120 seconds. \emph{Verified recovery} requires a ready, non-deadlocked board on the original level within 300 seconds, and a \emph{same-level win} requires the success event within the same limit. Every D continuation stays in the denominator, including runs that never recover. The 600-action ceiling never binds (the longest continuation has 24 actions); all 72 D continuations end at the time limit, and N has 19 completions and five timeouts, with no restart or voluntary stop.

\noindent\textbf{Estimation.} Table~\ref{tab:rq2-outcomes} reports the outcomes. Rates weight checkpoints equally within each state class, and paired effects compare outcomes at the same checkpoint. The 95\% percentile intervals resample the seven D source levels (10,000 bootstrap draws), keeping all checkpoints and conditions of each sampled level; they describe variation across sources, not repeat variability. Two serving endpoints with the same model and settings served the runs; on the 21 A/B pairs that share an endpoint, timely restarts rise from 1/21 to 19/21 and verified recovery from 8/21 to 20/21, and on the 23 B/C pairs that share one, verified recovery is 22/23 in both conditions.

\begin{table}[ht]
\centering
\small
\setlength{\tabcolsep}{4pt}
\caption{RQ2 outcomes for GPT-5.6-Sol. Each cell gives a count and percentage over all continuations in that state and condition. Timely restart must occur within 120 seconds and the first ten actions; verified recovery and same-level wins use the 300-second limit. No N continuation restarts, including after the response window.}
\label{tab:rq2-outcomes}
\begin{tabular}{@{}llrrr@{}}
\toprule
State & Condition & Timely restart & Verified recovery & Same-level win \\
\midrule
D & A & 1/24 (4.2\%) & 8/24 (33.3\%) & 0/24 (0.0\%) \\
D & B & 22/24 (91.7\%) & 23/24 (95.8\%) & 0/24 (0.0\%) \\
D & C & 22/24 (91.7\%) & 22/24 (91.7\%) & 0/24 (0.0\%) \\
\midrule
N & A & 0/12 (0.0\%) & --- & 10/12 (83.3\%) \\
N & B & 0/12 (0.0\%) & --- & 9/12 (75.0\%) \\
\bottomrule
\end{tabular}

\end{table}

\begin{table}[ht]
\centering
\small
\setlength{\tabcolsep}{5pt}
\caption{Within-level effects of state information on D, in percentage points. Every source level improves on timely selection and verified recovery. Checkpoints sharing a level remain together in the paired bootstrap. Win counts use the listed checkpoint count as the denominator for each condition.}
\label{tab:rq2-source-effects}
\begin{tabular}{@{}lrrrr@{}}
\toprule
Hard level & D checkpoints & Selection $B-A$ & Recovery $B-A$ & Wins (A/B/C) \\
\midrule
1 & 10 & +80.0 & +50.0 & 0/0/0 \\
2 & 2 & +100.0 & +100.0 & 0/0/0 \\
3 & 2 & +100.0 & +50.0 & 0/0/0 \\
5 & 2 & +100.0 & +50.0 & 0/0/0 \\
6 & 4 & +75.0 & +75.0 & 0/0/0 \\
7 & 2 & +100.0 & +50.0 & 0/0/0 \\
8 & 2 & +100.0 & +100.0 & 0/0/0 \\
\bottomrule
\end{tabular}

\end{table}

\noindent\textbf{Paired effects.} Timely restarts improve on 21 of the 24 paired D checkpoints and decline on none; verified recovery improves on 15 and declines on none. Seven of the eight recoveries under A occur after the response window, against one of 23 under B and none of 22 under C. From B to C, timely restarts improve on one checkpoint and decline on one (net zero; 95\% interval $[-7.9,15.0]$ points), and verified recovery differs on one checkpoint, the B/C pair that spans endpoints ($-4.2$ points; $[-7.9,0.0]$). On N, eight checkpoints finish under both A and B, one only under B, two only under A, and one under neither, and no run restarts.

\noindent\textbf{Attribution coding.} A D run \emph{blames the click} if any rationale describes a move as not taking effect or proposes retrying another target (A: 16/24, B: 3/24, C: 5/24). An unaided recovery \emph{names the cause} if a rationale before the restart states that the holding holes are full or that no destination remains (7 of the 8 recoveries under A).

\noindent\textbf{After recovery.} A recurrence is a new stable certificate on the target level after the first verified return to play and within 300 seconds. Of the 45 recovered B/C continuations, 34 recur (17 under each condition), 29 of them after positive progress from the restored board. The median recurrence comes 128.8 seconds after recovery, after eight accepted actions, with 148.5 seconds remaining. The 11 continuations without a new certificate also remain unfinished, despite a median 290.9 seconds available after recovery.

\noindent\textbf{Follow-up across models and games.} The follow-up repeats A/B/C with Sol, Terra, and Luna on six certified Bolt Unscrew dead ends and three solvable controls per model, and on three certified Nut and Bolt dead ends per model: 72 Bolt Unscrew and 27 Nut and Bolt trials (Table~\ref{tab:rq2-followup}). Pooled over Bolt Unscrew dead ends, verified recovery rises from 7/18 under A to 16/18 under B and stays at 16/18 under C; the effect is largest for Luna and nearly absent for Terra, which already recovers in 5/6 baseline trials. None of the 18 solvable controls selects recovery. In Nut and Bolt, recovery selection, verified recovery, and completion are all 0/27. All 81 follow-up dead-end continuations remain incomplete; with the original 72, this gives 0/153 same-level completions after dead-end takeover.

\begin{table}[ht]
\centering
\small
\setlength{\tabcolsep}{4pt}
\caption{Verified recovery in the RQ2 follow-up. Each Bolt Unscrew cell has six certified dead ends per condition; Nut and Bolt pools three dead ends per model, nine trials per condition. Same-level completion is zero in every cell.}
\label{tab:rq2-followup}
\begin{tabular}{@{}llccc@{}}
\toprule
Game & Model & A & B & C \\
\midrule
Bolt Unscrew & Luna & 0/6 & 5/6 & 5/6 \\
Bolt Unscrew & Sol & 2/6 & 5/6 & 6/6 \\
Bolt Unscrew & Terra & 5/6 & 6/6 & 5/6 \\
Nut and Bolt & All three & 0/9 & 0/9 & 0/9 \\
\bottomrule
\end{tabular}
\end{table}

\subsection{RQ3: matched retries with memory}
\label{sec:rq3-protocol}
\noindent\textbf{Design.} RQ3 branches from agent-owned Bolt Unscrew Hard failures after a verified native restart. Every branch starts from the same ready board, screenshot, objective, action interface, and budget. M0 clears the pre-restart history; M1 keeps the native history and its restart summary; M2 has the same capacity as M1 but replaces the restart summary with a structured, agent-written account of prior actions and outcomes, suspected failure causes, the restored state, and the proposed next strategy. After the restart, every condition accumulates history normally. The original Sol panel crosses 24 source failures with M0--M2 (72 retries); the follow-up crosses six fixed source failures with the three conditions and Sol, Terra, and Luna (54 retries). Fifteen model--source--condition units have one extra valid retry, used only to test endpoint stability.

\noindent\textbf{Endpoints.} A retry ends at same-level completion, recurrence of the source's certified failure, a different certified failure, another restart, an explicit stop, or the 300-second limit. Same-level completion is the primary outcome. Recurrence requires the same certified mechanism, not merely a similar board.

\noindent\textbf{Results.} The original panel yields 0/24 completions in every condition, and the follow-up 0/18 in M0, M1, and M2. In the follow-up, source-failure recurrence is 12/18, 16/18, and 15/18, and mean best progress is 0.124, 0.089, and 0.130. Paired within model and source, M2$-$M1 progress averages $+0.041$ but ranges from $-0.091$ to $+0.409$, and M2$-$M0 averages $+0.006$. Of the 15 repeated units, seven keep their endpoint and eight change it. The data thus support a null effect on completion and unstable endpoints, not a ranking of the memory conditions.

\noindent\textbf{Failure summaries.} Each of the 24 M2 summaries of the original panel has a ``Failure Experience'' and a ``Next Attempt'' section. A summary \emph{blames execution or observation} if the former cites unconfirmed outcomes, missing screenshots, or rejected placements (24/24); it \emph{resumes the failed plan} if the latter directs continuing or re-testing the previous intention (15/24); and it \emph{proposes a different plan} if the latter calls for an alternative or safer order (0/24). A manual reading confirms that no summary identifies exhaustion of the free holes as the cause. Source histories often end before the final dead-end frame, so a summary does not always observe the deadlock itself, although earlier frames in several histories already show the holding holes occupied.

\subsection{RQ4: rule card and workflow control}
\label{sec:mechanism-analyses}
\noindent\textbf{Design.} RQ4 varies two burdens inside complete sessions. Under K0, the agent must infer the mechanics through play; K1 adds a fixed game-level card with the legal interactions and constraints but no level-specific solution, coordinates, or action sequence. Under P0, the agent handles game entry, level advancement, and puzzle moves; P1 uses a deterministic controller for entry and evaluator-verified advancement and leaves every puzzle move to the agent. The original Sol study has 68 sessions: the four K/P cells on all 16 objectives, plus two sessions per Bolt Unscrew objective with a controller that restarts after a certified deadlock. Five isolated rule probes per session give 340 probe outcomes; probe responses never enter the acting history. The follow-up runs the four K/P cells with Sol, Terra, and Luna on six objectives (Bolt Unscrew Easy and Hard, Color Connect Easy, Maze Paint Easy and Hard, and Rush Hour Easy), for 72 sessions.

\begin{figure}[ht]
\centering
\includegraphics[width=\linewidth]{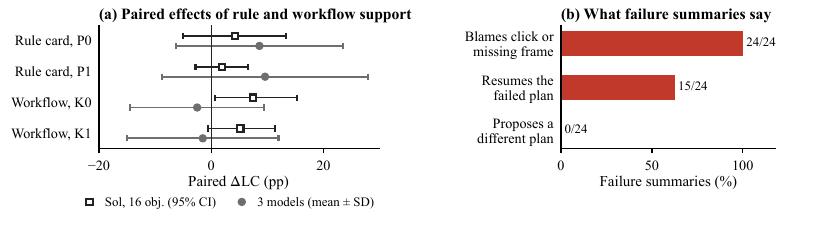}
\vspace{-14pt}
\caption{Supplementary views for RQ3--RQ4. (a) Paired LC effects of the rule card and workflow control at each level of the other factor: original Sol factorial over 16 objectives (bootstrap 95\% interval) and the three-model follow-up over 18 model--objective units (mean $\pm$ SD). (b) Content of the 24 agent-written failure summaries (M2).}
\label{fig:section5-support-details}
\end{figure}

\noindent\textbf{Contrasts.} LC is the primary outcome. For metric $Y$, the matched effects are
\begin{align}
\Delta_{K\mid P0}^{Y} &= Y_{K1P0}-Y_{K0P0}, &
\Delta_{K\mid P1}^{Y} &= Y_{K1P1}-Y_{K0P1}, \\
\Delta_{P\mid K0}^{Y} &= Y_{K0P1}-Y_{K0P0}, &
\Delta_{P\mid K1}^{Y} &= Y_{K1P1}-Y_{K1P0}, \\
\Delta_{K\times P}^{Y} &= \Delta_{K\mid P1}^{Y}-\Delta_{K\mid P0}^{Y}. &&
\end{align}
The original study pairs conditions within objective and reports an objective-within-game bootstrap interval; the follow-up computes each contrast within model and objective and reports the mean, standard deviation, and range over the 18 matched units (Table~\ref{tab:rq4-followup}, Figure~\ref{fig:section5-support-details}a).

\begin{table}[ht]
\centering
\small
\setlength{\tabcolsep}{4pt}
\caption{Cross-model RQ4 outcomes over 18 matched model--objective units per K/P cell. LC and FGP entries are mean paired effects in percentage points.}
\label{tab:rq4-followup}
\begin{tabular}{@{}lccc@{}}
\toprule
Contrast or cell & Completion & $\Delta$LC & $\Delta$FGP \\
\midrule
K0P0 & 5/18 & -- & -- \\
K0P1 & 6/18 & -- & -- \\
K1P0 & 8/18 & -- & -- \\
K1P1 & 8/18 & -- & -- \\
K1--K0 under P0 & -- & $+8.6$ & $+7.8$ \\
K1--K0 under P1 & -- & $+9.6$ & $+9.7$ \\
P1--P0 under K0 & -- & $-2.5$ & $-2.5$ \\
P1--P0 under K1 & -- & $-1.5$ & $-0.6$ \\
\bottomrule
\end{tabular}
\end{table}

\noindent\textbf{Rule knowledge.} In the original study, the first probe states 13.9\% and 16.1\% of rules correctly under K0P0 and K0P1, against 77.1\% and 86.3\% under K1P0 and K1P1. Figure~\ref{fig:section5-supports}c uses the final probe and LC of each P0 session, averaged over the 16 objectives within a K cell; under P1, the card raises final rule correctness from 52.3\% to 81.3\% and LC from 62.5\% to 65.4\%. These unweighted session means differ from the bootstrap contrasts of Figure~\ref{fig:section5-support-details}a. Without the card, 29 of 36 (P0) and 40 of 48 (P1) rules that are ever stated correctly are first stated correctly after the initial probe. Of the 40 sessions that state every rule correctly at some probe, 34 subsequently increase LC, and 16 still state every rule correctly at the final probe. The follow-up measures completion only.

\noindent\textbf{Workflow control.} P0 involves no controller actions. Across the 36 P1 follow-up sessions, the controller executes 111 entry and 164 advancement actions without error. Relative to P0, P1 reduces model actions by 29.4 per session under K0 and 19.1 under K1, and total actions, including the controller's, by 22.4 and 10.8. The K$\times$P interaction on LC is $+1.0$ points (standard deviation 19.3, range $[-40,+40]$), so the controller transfers workflow without a uniform gain in completion.

\clearpage
\section{Complete Per-Objective Results}
\label{sec:expanded-results}
Table~\ref{tab:gpt-denominators} gives the counts behind the aggregate metrics of Table~\ref{tab:main-results}, and Table~\ref{tab:gpt-episode-results} lists the outcome of every objective for every agent and setting. Every completion metric in Table~\ref{tab:main-results} follows from Table~\ref{tab:gpt-episode-results} by the game-macro average of Appendix~\ref{sec:metrics}; unvisited target levels contribute zero to LC and FGP. Values are rounded to one decimal place.

\begingroup\scriptsize
\setlength{\tabcolsep}{2pt}
\setlength{\LTleft}{\fill}\setlength{\LTright}{\fill}
\setlength{\LTcapwidth}{\textwidth}
\begin{longtable}{@{}lrrrrrrr@{}}
\caption{Metric denominators by evaluation setting. PER and SMR use all $n$ episodes. S is the number of successful episodes. Native counts visual-observation and GUI-input steps and Other all remaining model action steps (Code Execution CUA only). Cells: objectives with a valid episode.}\label{tab:gpt-denominators}\\
\toprule
Configuration & $n$ & S & PER & SMR & Native & Other & Cells \\
\midrule\endfirsthead
\multicolumn{8}{l}{\textit{Table~\ref{tab:gpt-denominators} continued}} \\
\toprule
Configuration & $n$ & S & PER & SMR & Native & Other & Cells \\
\midrule\endhead
\midrule\endfoot\bottomrule\endlastfoot
GPT-6-Astra / Code Execution CUA & 16 & 15 & 0 & 0 & 729 & 381 & 16/16 \\
GPT-6-Sol / Code Execution CUA & 16 & 13 & 3 & 0 & 486 & 696 & 16/16 \\
GPT-6-Luna / Code Execution CUA & 16 & 4 & 12 & 0 & 1620 & 858 & 16/16 \\
GPT-5.6-Sol / Code Execution CUA & 16 & 14 & 0 & 0 & 122 & 1316 & 16/16 \\
GPT-5.6-Terra / Code Execution CUA & 16 & 6 & 10 & 0 & 528 & 1067 & 16/16 \\
GPT-5.6-Luna / Code Execution CUA & 16 & 5 & 11 & 0 & 629 & 568 & 16/16 \\
DeepSeek 4.1 Flash / Code Execution CUA & 16 & 9 & 1 & 1 & 305 & 2303 & 16/16 \\
Qwen3.8 Flash / Code Execution CUA & 16 & 11 & 0 & 0 & 275 & 1295 & 16/16 \\
GLM-5.3-Flash / Code Execution CUA & 16 & 5 & 0 & 0 & 1324 & 1273 & 16/16 \\
Kimi-k3 / Code Execution CUA & 16 & 10 & 0 & 0 & 895 & 614 & 16/16 \\
GPT-6-Astra / Native GUI Actions & 16 & 15 & 0 & 0 & N/A & N/A & 16/16 \\
GPT-6-Sol / Native GUI Actions & 16 & 13 & 0 & 0 & N/A & N/A & 16/16 \\
GPT-6-Luna / Native GUI Actions & 16 & 4 & 11 & 0 & N/A & N/A & 16/16 \\
GPT-5.6-Sol / Native GUI Actions & 16 & 11 & 1 & 0 & N/A & N/A & 16/16 \\
GPT-5.6-Terra / Native GUI Actions & 16 & 7 & 9 & 0 & N/A & N/A & 16/16 \\
GPT-5.6-Luna / Native GUI Actions & 16 & 6 & 10 & 0 & N/A & N/A & 16/16 \\
DeepSeek 4.1 Flash / Native GUI Actions & 16 & 8 & 2 & 0 & N/A & N/A & 16/16 \\
Qwen3.8 Flash / Native GUI Actions & 16 & 8 & 0 & 0 & N/A & N/A & 16/16 \\
GLM-5.3-Flash / Native GUI Actions & 16 & 4 & 2 & 0 & N/A & N/A & 16/16 \\
Kimi-k3 / Native GUI Actions & 16 & 8 & 0 & 0 & N/A & N/A & 16/16 \\
UI-TARS-1.5-7B / Native GUI Actions (autonomous harness) & 16 & 0 & 14 & 0 & N/A & N/A & 16/16 \\
Qwen3-VL-32B-Thinking / Native GUI Actions (autonomous harness) & 16 & 0 & 0 & 0 & N/A & N/A & 16/16 \\
Qwen3-VL-30B-A3B-Thinking / Native GUI Actions (autonomous harness) & 16 & 0 & 3 & 0 & N/A & N/A & 16/16 \\
Qwen3-VL-8B-Thinking / Native GUI Actions (autonomous harness) & 16 & 0 & 0 & 0 & N/A & N/A & 16/16 \\
\end{longtable}\endgroup

\begingroup\scriptsize
\setlength{\tabcolsep}{2pt}
\setlength{\LTleft}{\fill}\setlength{\LTright}{\fill}
\setlength{\LTcapwidth}{\textwidth}
\begin{longtable}{@{}lrrrrrrr@{}}
\caption{Per-objective results for every agent and setting. S: objective solved; P: premature exit; M: setting mismatch. LC, FGP, and NGAR are percentages; NGAR applies to Code Execution CUA only (--). $|L_o|$: number of target levels.}\label{tab:gpt-episode-results}\\
\toprule
Objective & S & LC & FGP & P & M & NGAR & $|L_o|$ \\
\midrule\endfirsthead
\multicolumn{8}{l}{\textit{Table~\ref{tab:gpt-episode-results} continued}} \\
\toprule
Objective & S & LC & FGP & P & M & NGAR & $|L_o|$ \\
\midrule\endhead
\midrule\endfoot\bottomrule\endlastfoot
\multicolumn{8}{l}{\textbf{GPT-6-Astra / Code Execution CUA}} \\
Bolt / easy & 1 & 100.0 & 100.0 & 0 & 0 & 100.0 & 8 \\
Bolt / hard & 0 & 50.0 & 60.7 & 0 & 0 & 65.2 & 8 \\
Color / easy & 1 & 100.0 & 100.0 & 0 & 0 & 38.6 & 10 \\
Color / hard & 1 & 100.0 & 100.0 & 0 & 0 & 25.0 & 10 \\
Maze / easy & 1 & 100.0 & 100.0 & 0 & 0 & 92.0 & 10 \\
Maze / hard & 1 & 100.0 & 100.0 & 0 & 0 & 20.0 & 10 \\
Maze / medium & 1 & 100.0 & 100.0 & 0 & 0 & 27.0 & 10 \\
Nut / easy & 1 & 100.0 & 100.0 & 0 & 0 & 100.0 & 3 \\
Nut / extreme & 1 & 100.0 & 100.0 & 0 & 0 & 28.9 & 3 \\
Nut / hard & 1 & 100.0 & 100.0 & 0 & 0 & 26.3 & 3 \\
Nut / medium & 1 & 100.0 & 100.0 & 0 & 0 & 41.2 & 3 \\
Nut / nightmare & 1 & 100.0 & 100.0 & 0 & 0 & 48.0 & 1 \\
Rush / easy & 1 & 100.0 & 100.0 & 0 & 0 & 98.1 & 10 \\
Rush / hard & 1 & 100.0 & 100.0 & 0 & 0 & 58.6 & 10 \\
Rush / medium & 1 & 100.0 & 100.0 & 0 & 0 & 85.7 & 10 \\
Truck / default & 1 & 100.0 & 100.0 & 0 & 0 & 100.0 & 5 \\
\multicolumn{8}{l}{\textbf{GPT-6-Sol / Code Execution CUA}} \\
Bolt / easy & 1 & 100.0 & 100.0 & 0 & 0 & 100.0 & 8 \\
Bolt / hard & 0 & 0.0 & 3.6 & 1 & 0 & 38.5 & 8 \\
Color / easy & 1 & 100.0 & 100.0 & 0 & 0 & 79.0 & 10 \\
Color / hard & 0 & 0.0 & 2.5 & 1 & 0 & 80.6 & 10 \\
Maze / easy & 1 & 100.0 & 100.0 & 0 & 0 & 30.4 & 10 \\
Maze / hard & 1 & 100.0 & 100.0 & 0 & 0 & 21.0 & 10 \\
Maze / medium & 1 & 100.0 & 100.0 & 0 & 0 & 8.8 & 10 \\
Nut / easy & 1 & 100.0 & 100.0 & 0 & 0 & 18.9 & 3 \\
Nut / extreme & 1 & 100.0 & 100.0 & 0 & 0 & 8.1 & 3 \\
Nut / hard & 0 & 66.7 & 91.4 & 1 & 0 & 24.6 & 3 \\
Nut / medium & 1 & 100.0 & 100.0 & 0 & 0 & 20.8 & 3 \\
Nut / nightmare & 1 & 100.0 & 100.0 & 0 & 0 & 11.6 & 1 \\
Rush / easy & 1 & 100.0 & 100.0 & 0 & 0 & 69.8 & 10 \\
Rush / hard & 1 & 100.0 & 100.0 & 0 & 0 & 78.3 & 10 \\
Rush / medium & 1 & 100.0 & 100.0 & 0 & 0 & 98.4 & 10 \\
Truck / default & 1 & 100.0 & 100.0 & 0 & 0 & 16.7 & 5 \\
\multicolumn{8}{l}{\textbf{GPT-6-Luna / Code Execution CUA}} \\
Bolt / easy & 1 & 100.0 & 100.0 & 0 & 0 & 98.5 & 8 \\
Bolt / hard & 0 & 0.0 & 3.6 & 1 & 0 & 92.9 & 8 \\
Color / easy & 0 & 50.0 & 56.7 & 1 & 0 & 81.4 & 10 \\
Color / hard & 0 & 40.0 & 40.0 & 1 & 0 & 89.3 & 10 \\
Maze / easy & 1 & 100.0 & 100.0 & 0 & 0 & 9.7 & 10 \\
Maze / hard & 0 & 20.0 & 25.8 & 1 & 0 & 3.8 & 10 \\
Maze / medium & 0 & 90.0 & 98.3 & 1 & 0 & 29.0 & 10 \\
Nut / easy & 0 & 0.0 & 25.0 & 1 & 0 & 98.8 & 3 \\
Nut / extreme & 0 & 0.0 & 3.2 & 1 & 0 & 27.3 & 3 \\
Nut / hard & 0 & 33.3 & 40.5 & 1 & 0 & 29.9 & 3 \\
Nut / medium & 0 & 33.3 & 33.3 & 1 & 0 & 31.7 & 3 \\
Nut / nightmare & 0 & 0.0 & 12.5 & 1 & 0 & 86.0 & 1 \\
Rush / easy & 1 & 100.0 & 100.0 & 0 & 0 & 92.9 & 10 \\
Rush / hard & 0 & 60.0 & 62.3 & 1 & 0 & 95.2 & 10 \\
Rush / medium & 0 & 40.0 & 46.7 & 1 & 0 & 97.2 & 10 \\
Truck / default & 1 & 100.0 & 100.0 & 0 & 0 & 19.7 & 5 \\
\multicolumn{8}{l}{\textbf{GPT-5.6-Sol / Code Execution CUA}} \\
Bolt / easy & 1 & 100.0 & 100.0 & 0 & 0 & 10.9 & 8 \\
Bolt / hard & 0 & 12.5 & 15.3 & 0 & 0 & 6.8 & 8 \\
Color / easy & 1 & 100.0 & 100.0 & 0 & 0 & 4.8 & 10 \\
Color / hard & 1 & 100.0 & 100.0 & 0 & 0 & 16.7 & 10 \\
Maze / easy & 1 & 100.0 & 100.0 & 0 & 0 & 2.4 & 10 \\
Maze / hard & 1 & 100.0 & 100.0 & 0 & 0 & 5.4 & 10 \\
Maze / medium & 1 & 100.0 & 100.0 & 0 & 0 & 3.2 & 10 \\
Nut / easy & 1 & 100.0 & 100.0 & 0 & 0 & 21.4 & 3 \\
Nut / extreme & 1 & 100.0 & 100.0 & 0 & 0 & 2.9 & 3 \\
Nut / hard & 0 & 66.7 & 69.8 & 0 & 0 & 1.1 & 3 \\
Nut / medium & 1 & 100.0 & 100.0 & 0 & 0 & 8.3 & 3 \\
Nut / nightmare & 1 & 100.0 & 100.0 & 0 & 0 & 6.5 & 1 \\
Rush / easy & 1 & 100.0 & 100.0 & 0 & 0 & 29.8 & 10 \\
Rush / hard & 1 & 100.0 & 100.0 & 0 & 0 & 50.0 & 10 \\
Rush / medium & 1 & 100.0 & 100.0 & 0 & 0 & 2.7 & 10 \\
Truck / default & 1 & 100.0 & 100.0 & 0 & 0 & 2.6 & 5 \\
\multicolumn{8}{l}{\textbf{GPT-5.6-Terra / Code Execution CUA}} \\
Bolt / easy & 0 & 87.5 & 98.6 & 1 & 0 & 18.6 & 8 \\
Bolt / hard & 0 & 0.0 & 5.2 & 1 & 0 & 18.2 & 8 \\
Color / easy & 0 & 70.0 & 78.3 & 1 & 0 & 77.3 & 10 \\
Color / hard & 1 & 100.0 & 100.0 & 0 & 0 & 23.2 & 10 \\
Maze / easy & 1 & 100.0 & 100.0 & 0 & 0 & 0.7 & 10 \\
Maze / hard & 0 & 30.0 & 39.5 & 1 & 0 & 3.9 & 10 \\
Maze / medium & 1 & 100.0 & 100.0 & 0 & 0 & 3.3 & 10 \\
Nut / easy & 1 & 100.0 & 100.0 & 0 & 0 & 2.6 & 3 \\
Nut / extreme & 0 & 0.0 & 5.1 & 1 & 0 & 10.8 & 3 \\
Nut / hard & 0 & 0.0 & 5.3 & 1 & 0 & 10.3 & 3 \\
Nut / medium & 0 & 33.3 & 38.7 & 1 & 0 & 2.5 & 3 \\
Nut / nightmare & 0 & 0.0 & 16.7 & 1 & 0 & 3.4 & 1 \\
Rush / easy & 1 & 100.0 & 100.0 & 0 & 0 & 98.0 & 10 \\
Rush / hard & 0 & 50.0 & 55.0 & 1 & 0 & 85.3 & 10 \\
Rush / medium & 0 & 90.0 & 91.4 & 1 & 0 & 99.0 & 10 \\
Truck / default & 1 & 100.0 & 100.0 & 0 & 0 & 4.4 & 5 \\
\multicolumn{8}{l}{\textbf{GPT-5.6-Luna / Code Execution CUA}} \\
Bolt / easy & 1 & 100.0 & 100.0 & 0 & 0 & 2.8 & 8 \\
Bolt / hard & 0 & 0.0 & 3.6 & 1 & 0 & 88.5 & 8 \\
Color / easy & 0 & 0.0 & 6.7 & 1 & 0 & 56.0 & 10 \\
Color / hard & 0 & 0.0 & 8.8 & 1 & 0 & 77.8 & 10 \\
Maze / easy & 1 & 100.0 & 100.0 & 0 & 0 & 95.2 & 10 \\
Maze / hard & 0 & 50.0 & 58.6 & 1 & 0 & 9.2 & 10 \\
Maze / medium & 0 & 0.0 & 9.3 & 1 & 0 & 12.0 & 10 \\
Nut / easy & 1 & 100.0 & 100.0 & 0 & 0 & 10.6 & 3 \\
Nut / extreme & 0 & 0.0 & 1.9 & 1 & 0 & 36.6 & 3 \\
Nut / hard & 0 & 0.0 & 2.3 & 1 & 0 & 43.5 & 3 \\
Nut / medium & 0 & 33.3 & 37.3 & 1 & 0 & 33.3 & 3 \\
Nut / nightmare & 0 & 0.0 & 15.3 & 1 & 0 & 45.2 & 1 \\
Rush / easy & 1 & 100.0 & 100.0 & 0 & 0 & 63.0 & 10 \\
Rush / hard & 0 & 0.0 & 6.4 & 1 & 0 & 88.5 & 10 \\
Rush / medium & 0 & 40.0 & 45.5 & 1 & 0 & 74.3 & 10 \\
Truck / default & 1 & 100.0 & 100.0 & 0 & 0 & 0.9 & 5 \\
\multicolumn{8}{l}{\textbf{DeepSeek 4.1 Flash / Code Execution CUA}} \\
Bolt / easy & 1 & 100.0 & 100.0 & 0 & 0 & 3.0 & 8 \\
Bolt / hard & 0 & 12.5 & 17.6 & 1 & 1 & 66.4 & 8 \\
Color / easy & 1 & 100.0 & 100.0 & 0 & 0 & 2.7 & 10 \\
Color / hard & 1 & 100.0 & 100.0 & 0 & 0 & 5.7 & 10 \\
Maze / easy & 1 & 100.0 & 100.0 & 0 & 0 & 1.6 & 10 \\
Maze / hard & 0 & 40.0 & 49.8 & 0 & 0 & 13.2 & 10 \\
Maze / medium & 1 & 100.0 & 100.0 & 0 & 0 & 1.9 & 10 \\
Nut / easy & 1 & 100.0 & 100.0 & 0 & 0 & 4.2 & 3 \\
Nut / extreme & 0 & 0.0 & 5.1 & 0 & 0 & 2.1 & 3 \\
Nut / hard & 0 & 0.0 & 9.1 & 0 & 0 & 1.2 & 3 \\
Nut / medium & 0 & 66.7 & 73.0 & 0 & 0 & 0.9 & 3 \\
Nut / nightmare & 0 & 0.0 & 18.1 & 0 & 0 & 5.2 & 1 \\
Rush / easy & 1 & 100.0 & 100.0 & 0 & 0 & 1.9 & 10 \\
Rush / hard & 0 & 70.0 & 76.5 & 0 & 0 & 0.7 & 10 \\
Rush / medium & 1 & 100.0 & 100.0 & 0 & 0 & 64.8 & 10 \\
Truck / default & 1 & 100.0 & 100.0 & 0 & 0 & 1.9 & 5 \\
\multicolumn{8}{l}{\textbf{Qwen3.8 Flash / Code Execution CUA}} \\
Bolt / easy & 1 & 100.0 & 100.0 & 0 & 0 & 3.3 & 8 \\
Bolt / hard & 0 & 12.5 & 15.1 & 0 & 0 & 3.9 & 8 \\
Color / easy & 1 & 100.0 & 100.0 & 0 & 0 & 17.9 & 10 \\
Color / hard & 1 & 100.0 & 100.0 & 0 & 0 & 14.9 & 10 \\
Maze / easy & 1 & 100.0 & 100.0 & 0 & 0 & 6.0 & 10 \\
Maze / hard & 0 & 50.0 & 58.8 & 0 & 0 & 5.3 & 10 \\
Maze / medium & 1 & 100.0 & 100.0 & 0 & 0 & 3.5 & 10 \\
Nut / easy & 1 & 100.0 & 100.0 & 0 & 0 & 42.5 & 3 \\
Nut / extreme & 0 & 33.3 & 37.8 & 0 & 0 & 7.9 & 3 \\
Nut / hard & 0 & 0.0 & 6.1 & 0 & 0 & 10.2 & 3 \\
Nut / medium & 1 & 100.0 & 100.0 & 0 & 0 & 13.1 & 3 \\
Nut / nightmare & 0 & 0.0 & 15.3 & 0 & 0 & 28.6 & 1 \\
Rush / easy & 1 & 100.0 & 100.0 & 0 & 0 & 35.1 & 10 \\
Rush / hard & 1 & 100.0 & 100.0 & 0 & 0 & 37.9 & 10 \\
Rush / medium & 1 & 100.0 & 100.0 & 0 & 0 & 97.0 & 10 \\
Truck / default & 1 & 100.0 & 100.0 & 0 & 0 & 11.8 & 5 \\
\multicolumn{8}{l}{\textbf{GLM-5.3-Flash / Code Execution CUA}} \\
Bolt / easy & 1 & 100.0 & 100.0 & 0 & 0 & 1.9 & 8 \\
Bolt / hard & 0 & 12.5 & 24.4 & 0 & 0 & 3.0 & 8 \\
Color / easy & 0 & 20.0 & 29.1 & 0 & 0 & 33.3 & 10 \\
Color / hard & 0 & 20.0 & 20.0 & 0 & 0 & 6.4 & 10 \\
Maze / easy & 1 & 100.0 & 100.0 & 0 & 0 & 4.3 & 10 \\
Maze / hard & 0 & 10.0 & 18.3 & 0 & 0 & 100.0 & 10 \\
Maze / medium & 0 & 60.0 & 68.4 & 0 & 0 & 0.8 & 10 \\
Nut / easy & 1 & 100.0 & 100.0 & 0 & 0 & 98.9 & 3 \\
Nut / extreme & 0 & 0.0 & 6.4 & 0 & 0 & 7.8 & 3 \\
Nut / hard & 0 & 0.0 & 7.6 & 0 & 0 & 3.0 & 3 \\
Nut / medium & 0 & 33.3 & 56.0 & 0 & 0 & 1.0 & 3 \\
Nut / nightmare & 0 & 0.0 & 13.9 & 0 & 0 & 100.0 & 1 \\
Rush / easy & 1 & 100.0 & 100.0 & 0 & 0 & 88.7 & 10 \\
Rush / hard & 0 & 30.0 & 37.6 & 0 & 0 & 0.6 & 10 \\
Rush / medium & 0 & 30.0 & 32.7 & 0 & 0 & 100.0 & 10 \\
Truck / default & 1 & 100.0 & 100.0 & 0 & 0 & 99.6 & 5 \\
\multicolumn{8}{l}{\textbf{Kimi-k3 / Code Execution CUA}} \\
Bolt / easy & 1 & 100.0 & 100.0 & 0 & 0 & 98.7 & 8 \\
Bolt / hard & 0 & 25.0 & 26.1 & 0 & 0 & 100.0 & 8 \\
Color / easy & 1 & 100.0 & 100.0 & 0 & 0 & 95.1 & 10 \\
Color / hard & 1 & 100.0 & 100.0 & 0 & 0 & 2.3 & 10 \\
Maze / easy & 1 & 100.0 & 100.0 & 0 & 0 & 91.8 & 10 \\
Maze / hard & 0 & 30.0 & 33.1 & 0 & 0 & 11.7 & 10 \\
Maze / medium & 1 & 100.0 & 100.0 & 0 & 0 & 17.6 & 10 \\
Nut / easy & 1 & 100.0 & 100.0 & 0 & 0 & 14.8 & 3 \\
Nut / extreme & 0 & 0.0 & 5.1 & 0 & 0 & 14.9 & 3 \\
Nut / hard & 0 & 0.0 & 6.1 & 0 & 0 & 17.1 & 3 \\
Nut / medium & 1 & 100.0 & 100.0 & 0 & 0 & 2.4 & 3 \\
Nut / nightmare & 0 & 0.0 & 13.9 & 0 & 0 & 50.9 & 1 \\
Rush / easy & 1 & 100.0 & 100.0 & 0 & 0 & 99.2 & 10 \\
Rush / hard & 0 & 60.0 & 60.0 & 0 & 0 & 10.0 & 10 \\
Rush / medium & 1 & 100.0 & 100.0 & 0 & 0 & 16.1 & 10 \\
Truck / default & 1 & 100.0 & 100.0 & 0 & 0 & 76.5 & 5 \\
\multicolumn{8}{l}{\textbf{GPT-6-Astra / Native GUI Actions}} \\
Bolt / easy & 1 & 100.0 & 100.0 & 0 & 0 & -- & 8 \\
Bolt / hard & 0 & 37.5 & 37.9 & 0 & 0 & -- & 8 \\
Color / easy & 1 & 100.0 & 100.0 & 0 & 0 & -- & 10 \\
Color / hard & 1 & 100.0 & 100.0 & 0 & 0 & -- & 10 \\
Maze / easy & 1 & 100.0 & 100.0 & 0 & 0 & -- & 10 \\
Maze / hard & 1 & 100.0 & 100.0 & 0 & 0 & -- & 10 \\
Maze / medium & 1 & 100.0 & 100.0 & 0 & 0 & -- & 10 \\
Nut / easy & 1 & 100.0 & 100.0 & 0 & 0 & -- & 3 \\
Nut / extreme & 1 & 100.0 & 100.0 & 0 & 0 & -- & 3 \\
Nut / hard & 1 & 100.0 & 100.0 & 0 & 0 & -- & 3 \\
Nut / medium & 1 & 100.0 & 100.0 & 0 & 0 & -- & 3 \\
Nut / nightmare & 1 & 100.0 & 100.0 & 0 & 0 & -- & 1 \\
Rush / easy & 1 & 100.0 & 100.0 & 0 & 0 & -- & 10 \\
Rush / hard & 1 & 100.0 & 100.0 & 0 & 0 & -- & 10 \\
Rush / medium & 1 & 100.0 & 100.0 & 0 & 0 & -- & 10 \\
Truck / default & 1 & 100.0 & 100.0 & 0 & 0 & -- & 5 \\
\multicolumn{8}{l}{\textbf{GPT-6-Sol / Native GUI Actions}} \\
Bolt / easy & 1 & 100.0 & 100.0 & 0 & 0 & -- & 8 \\
Bolt / hard & 0 & 25.0 & 29.8 & 0 & 0 & -- & 8 \\
Color / easy & 1 & 100.0 & 100.0 & 0 & 0 & -- & 10 \\
Color / hard & 1 & 100.0 & 100.0 & 0 & 0 & -- & 10 \\
Maze / easy & 1 & 100.0 & 100.0 & 0 & 0 & -- & 10 \\
Maze / hard & 0 & 70.0 & 74.2 & 0 & 0 & -- & 10 \\
Maze / medium & 1 & 100.0 & 100.0 & 0 & 0 & -- & 10 \\
Nut / easy & 1 & 100.0 & 100.0 & 0 & 0 & -- & 3 \\
Nut / extreme & 1 & 100.0 & 100.0 & 0 & 0 & -- & 3 \\
Nut / hard & 0 & 66.7 & 84.0 & 0 & 0 & -- & 3 \\
Nut / medium & 1 & 100.0 & 100.0 & 0 & 0 & -- & 3 \\
Nut / nightmare & 1 & 100.0 & 100.0 & 0 & 0 & -- & 1 \\
Rush / easy & 1 & 100.0 & 100.0 & 0 & 0 & -- & 10 \\
Rush / hard & 1 & 100.0 & 100.0 & 0 & 0 & -- & 10 \\
Rush / medium & 1 & 100.0 & 100.0 & 0 & 0 & -- & 10 \\
Truck / default & 1 & 100.0 & 100.0 & 0 & 0 & -- & 5 \\
\multicolumn{8}{l}{\textbf{GPT-6-Luna / Native GUI Actions}} \\
Bolt / easy & 1 & 100.0 & 100.0 & 0 & 0 & -- & 8 \\
Bolt / hard & 0 & 0.0 & 4.4 & 1 & 0 & -- & 8 \\
Color / easy & 0 & 0.0 & 1.7 & 1 & 0 & -- & 10 \\
Color / hard & 0 & 0.0 & 7.5 & 1 & 0 & -- & 10 \\
Maze / easy & 1 & 100.0 & 100.0 & 0 & 0 & -- & 10 \\
Maze / hard & 0 & 0.0 & 1.8 & 1 & 0 & -- & 10 \\
Maze / medium & 0 & 50.0 & 57.5 & 1 & 0 & -- & 10 \\
Nut / easy & 1 & 100.0 & 100.0 & 0 & 0 & -- & 3 \\
Nut / extreme & 0 & 0.0 & 1.3 & 1 & 0 & -- & 3 \\
Nut / hard & 0 & 0.0 & 6.1 & 0 & 0 & -- & 3 \\
Nut / medium & 0 & 0.0 & 25.0 & 1 & 0 & -- & 3 \\
Nut / nightmare & 0 & 0.0 & 11.1 & 1 & 0 & -- & 1 \\
Rush / easy & 0 & 20.0 & 22.0 & 1 & 0 & -- & 10 \\
Rush / hard & 0 & 0.0 & 6.2 & 1 & 0 & -- & 10 \\
Rush / medium & 0 & 40.0 & 45.1 & 1 & 0 & -- & 10 \\
Truck / default & 1 & 100.0 & 100.0 & 0 & 0 & -- & 5 \\
\multicolumn{8}{l}{\textbf{GPT-5.6-Sol / Native GUI Actions}} \\
Bolt / easy & 1 & 100.0 & 100.0 & 0 & 0 & -- & 8 \\
Bolt / hard & 0 & 25.0 & 26.8 & 0 & 0 & -- & 8 \\
Color / easy & 1 & 100.0 & 100.0 & 0 & 0 & -- & 10 \\
Color / hard & 0 & 20.0 & 29.0 & 1 & 0 & -- & 10 \\
Maze / easy & 1 & 100.0 & 100.0 & 0 & 0 & -- & 10 \\
Maze / hard & 0 & 90.0 & 98.4 & 0 & 0 & -- & 10 \\
Maze / medium & 1 & 100.0 & 100.0 & 0 & 0 & -- & 10 \\
Nut / easy & 1 & 100.0 & 100.0 & 0 & 0 & -- & 3 \\
Nut / extreme & 0 & 66.7 & 69.4 & 0 & 0 & -- & 3 \\
Nut / hard & 0 & 66.7 & 74.1 & 0 & 0 & -- & 3 \\
Nut / medium & 1 & 100.0 & 100.0 & 0 & 0 & -- & 3 \\
Nut / nightmare & 1 & 100.0 & 100.0 & 0 & 0 & -- & 1 \\
Rush / easy & 1 & 100.0 & 100.0 & 0 & 0 & -- & 10 \\
Rush / hard & 1 & 100.0 & 100.0 & 0 & 0 & -- & 10 \\
Rush / medium & 1 & 100.0 & 100.0 & 0 & 0 & -- & 10 \\
Truck / default & 1 & 100.0 & 100.0 & 0 & 0 & -- & 5 \\
\multicolumn{8}{l}{\textbf{GPT-5.6-Terra / Native GUI Actions}} \\
Bolt / easy & 1 & 100.0 & 100.0 & 0 & 0 & -- & 8 \\
Bolt / hard & 0 & 12.5 & 17.9 & 1 & 0 & -- & 8 \\
Color / easy & 0 & 50.0 & 56.7 & 1 & 0 & -- & 10 \\
Color / hard & 0 & 20.0 & 28.0 & 1 & 0 & -- & 10 \\
Maze / easy & 1 & 100.0 & 100.0 & 0 & 0 & -- & 10 \\
Maze / hard & 0 & 30.0 & 34.8 & 1 & 0 & -- & 10 \\
Maze / medium & 0 & 40.0 & 46.9 & 1 & 0 & -- & 10 \\
Nut / easy & 1 & 100.0 & 100.0 & 0 & 0 & -- & 3 \\
Nut / extreme & 0 & 0.0 & 9.0 & 1 & 0 & -- & 3 \\
Nut / hard & 0 & 0.0 & 6.1 & 1 & 0 & -- & 3 \\
Nut / medium & 1 & 100.0 & 100.0 & 0 & 0 & -- & 3 \\
Nut / nightmare & 0 & 0.0 & 13.9 & 1 & 0 & -- & 1 \\
Rush / easy & 1 & 100.0 & 100.0 & 0 & 0 & -- & 10 \\
Rush / hard & 0 & 40.0 & 42.2 & 1 & 0 & -- & 10 \\
Rush / medium & 1 & 100.0 & 100.0 & 0 & 0 & -- & 10 \\
Truck / default & 1 & 100.0 & 100.0 & 0 & 0 & -- & 5 \\
\multicolumn{8}{l}{\textbf{GPT-5.6-Luna / Native GUI Actions}} \\
Bolt / easy & 1 & 100.0 & 100.0 & 0 & 0 & -- & 8 \\
Bolt / hard & 0 & 0.0 & 3.6 & 1 & 0 & -- & 8 \\
Color / easy & 0 & 60.0 & 60.0 & 1 & 0 & -- & 10 \\
Color / hard & 0 & 0.0 & 5.0 & 1 & 0 & -- & 10 \\
Maze / easy & 1 & 100.0 & 100.0 & 0 & 0 & -- & 10 \\
Maze / hard & 0 & 0.0 & 6.4 & 1 & 0 & -- & 10 \\
Maze / medium & 1 & 100.0 & 100.0 & 0 & 0 & -- & 10 \\
Nut / easy & 1 & 100.0 & 100.0 & 0 & 0 & -- & 3 \\
Nut / extreme & 0 & 0.0 & 5.1 & 1 & 0 & -- & 3 \\
Nut / hard & 0 & 0.0 & 6.1 & 1 & 0 & -- & 3 \\
Nut / medium & 0 & 0.0 & 23.6 & 1 & 0 & -- & 3 \\
Nut / nightmare & 0 & 0.0 & 9.7 & 1 & 0 & -- & 1 \\
Rush / easy & 1 & 100.0 & 100.0 & 0 & 0 & -- & 10 \\
Rush / hard & 0 & 20.0 & 22.7 & 1 & 0 & -- & 10 \\
Rush / medium & 0 & 40.0 & 45.9 & 1 & 0 & -- & 10 \\
Truck / default & 1 & 100.0 & 100.0 & 0 & 0 & -- & 5 \\
\multicolumn{8}{l}{\textbf{DeepSeek 4.1 Flash / Native GUI Actions}} \\
Bolt / easy & 0 & 87.5 & 95.8 & 0 & 0 & -- & 8 \\
Bolt / hard & 0 & 0.0 & 4.2 & 1 & 0 & -- & 8 \\
Color / easy & 1 & 100.0 & 100.0 & 0 & 0 & -- & 10 \\
Color / hard & 0 & 70.0 & 72.7 & 0 & 0 & -- & 10 \\
Maze / easy & 1 & 100.0 & 100.0 & 0 & 0 & -- & 10 \\
Maze / hard & 0 & 50.0 & 59.3 & 0 & 0 & -- & 10 \\
Maze / medium & 1 & 100.0 & 100.0 & 0 & 0 & -- & 10 \\
Nut / easy & 1 & 100.0 & 100.0 & 0 & 0 & -- & 3 \\
Nut / extreme & 0 & 0.0 & 6.4 & 0 & 0 & -- & 3 \\
Nut / hard & 0 & 0.0 & 6.1 & 0 & 0 & -- & 3 \\
Nut / medium & 1 & 100.0 & 100.0 & 0 & 0 & -- & 3 \\
Nut / nightmare & 0 & 0.0 & 15.3 & 0 & 0 & -- & 1 \\
Rush / easy & 1 & 100.0 & 100.0 & 0 & 0 & -- & 10 \\
Rush / hard & 0 & 40.0 & 42.9 & 1 & 0 & -- & 10 \\
Rush / medium & 1 & 100.0 & 100.0 & 0 & 0 & -- & 10 \\
Truck / default & 1 & 100.0 & 100.0 & 0 & 0 & -- & 5 \\
\multicolumn{8}{l}{\textbf{Qwen3.8 Flash / Native GUI Actions}} \\
Bolt / easy & 1 & 100.0 & 100.0 & 0 & 0 & -- & 8 \\
Bolt / hard & 0 & 0.0 & 5.7 & 0 & 0 & -- & 8 \\
Color / easy & 0 & 90.0 & 98.0 & 0 & 0 & -- & 10 \\
Color / hard & 0 & 30.0 & 38.3 & 0 & 0 & -- & 10 \\
Maze / easy & 1 & 100.0 & 100.0 & 0 & 0 & -- & 10 \\
Maze / hard & 0 & 30.0 & 36.7 & 0 & 0 & -- & 10 \\
Maze / medium & 1 & 100.0 & 100.0 & 0 & 0 & -- & 10 \\
Nut / easy & 1 & 100.0 & 100.0 & 0 & 0 & -- & 3 \\
Nut / extreme & 0 & 0.0 & 23.1 & 0 & 0 & -- & 3 \\
Nut / hard & 0 & 0.0 & 14.4 & 0 & 0 & -- & 3 \\
Nut / medium & 1 & 100.0 & 100.0 & 0 & 0 & -- & 3 \\
Nut / nightmare & 0 & 0.0 & 19.4 & 0 & 0 & -- & 1 \\
Rush / easy & 1 & 100.0 & 100.0 & 0 & 0 & -- & 10 \\
Rush / hard & 0 & 30.0 & 33.3 & 0 & 0 & -- & 10 \\
Rush / medium & 1 & 100.0 & 100.0 & 0 & 0 & -- & 10 \\
Truck / default & 1 & 100.0 & 100.0 & 0 & 0 & -- & 5 \\
\multicolumn{8}{l}{\textbf{GLM-5.3-Flash / Native GUI Actions}} \\
Bolt / easy & 1 & 100.0 & 100.0 & 0 & 0 & -- & 8 \\
Bolt / hard & 0 & 0.0 & 0.5 & 1 & 0 & -- & 8 \\
Color / easy & 0 & 90.0 & 90.0 & 0 & 0 & -- & 10 \\
Color / hard & 0 & 0.0 & 5.0 & 1 & 0 & -- & 10 \\
Maze / easy & 0 & 60.0 & 67.0 & 0 & 0 & -- & 10 \\
Maze / hard & 0 & 20.0 & 28.3 & 0 & 0 & -- & 10 \\
Maze / medium & 0 & 50.0 & 52.2 & 0 & 0 & -- & 10 \\
Nut / easy & 1 & 100.0 & 100.0 & 0 & 0 & -- & 3 \\
Nut / extreme & 0 & 0.0 & 5.1 & 0 & 0 & -- & 3 \\
Nut / hard & 0 & 0.0 & 6.8 & 0 & 0 & -- & 3 \\
Nut / medium & 0 & 33.3 & 40.0 & 0 & 0 & -- & 3 \\
Nut / nightmare & 0 & 0.0 & 16.7 & 0 & 0 & -- & 1 \\
Rush / easy & 1 & 100.0 & 100.0 & 0 & 0 & -- & 10 \\
Rush / hard & 0 & 0.0 & 6.4 & 0 & 0 & -- & 10 \\
Rush / medium & 0 & 40.0 & 45.9 & 0 & 0 & -- & 10 \\
Truck / default & 1 & 100.0 & 100.0 & 0 & 0 & -- & 5 \\
\multicolumn{8}{l}{\textbf{Kimi-k3 / Native GUI Actions}} \\
Bolt / easy & 1 & 100.0 & 100.0 & 0 & 0 & -- & 8 \\
Bolt / hard & 0 & 0.0 & 5.7 & 0 & 0 & -- & 8 \\
Color / easy & 1 & 100.0 & 100.0 & 0 & 0 & -- & 10 \\
Color / hard & 0 & 40.0 & 48.3 & 0 & 0 & -- & 10 \\
Maze / easy & 1 & 100.0 & 100.0 & 0 & 0 & -- & 10 \\
Maze / hard & 0 & 30.0 & 30.0 & 0 & 0 & -- & 10 \\
Maze / medium & 0 & 30.0 & 37.9 & 0 & 0 & -- & 10 \\
Nut / easy & 1 & 100.0 & 100.0 & 0 & 0 & -- & 3 \\
Nut / extreme & 0 & 0.0 & 5.8 & 0 & 0 & -- & 3 \\
Nut / hard & 0 & 33.3 & 33.3 & 0 & 0 & -- & 3 \\
Nut / medium & 1 & 100.0 & 100.0 & 0 & 0 & -- & 3 \\
Nut / nightmare & 0 & 0.0 & 19.4 & 0 & 0 & -- & 1 \\
Rush / easy & 1 & 100.0 & 100.0 & 0 & 0 & -- & 10 \\
Rush / hard & 0 & 0.0 & 6.2 & 0 & 0 & -- & 10 \\
Rush / medium & 1 & 100.0 & 100.0 & 0 & 0 & -- & 10 \\
Truck / default & 1 & 100.0 & 100.0 & 0 & 0 & -- & 5 \\
\multicolumn{8}{l}{\textbf{UI-TARS-1.5-7B / Native GUI Actions (autonomous harness)}} \\
Bolt / easy & 0 & 12.5 & 12.5 & 1 & 0 & -- & 8 \\
Bolt / hard & 0 & 0.0 & 0.0 & 1 & 0 & -- & 8 \\
Color / easy & 0 & 0.0 & 0.0 & 1 & 0 & -- & 10 \\
Color / hard & 0 & 0.0 & 0.0 & 1 & 0 & -- & 10 \\
Maze / easy & 0 & 0.0 & 0.0 & 1 & 0 & -- & 10 \\
Maze / hard & 0 & 0.0 & 0.0 & 1 & 0 & -- & 10 \\
Maze / medium & 0 & 0.0 & 7.1 & 1 & 0 & -- & 10 \\
Nut / easy & 0 & 33.3 & 51.9 & 1 & 0 & -- & 3 \\
Nut / extreme & 0 & 0.0 & 0.0 & 0 & 0 & -- & 3 \\
Nut / hard & 0 & 0.0 & 0.0 & 0 & 0 & -- & 3 \\
Nut / medium & 0 & 0.0 & 4.2 & 1 & 0 & -- & 3 \\
Nut / nightmare & 0 & 0.0 & 0.0 & 1 & 0 & -- & 1 \\
Rush / easy & 0 & 0.0 & 0.0 & 1 & 0 & -- & 10 \\
Rush / hard & 0 & 0.0 & 0.0 & 1 & 0 & -- & 10 \\
Rush / medium & 0 & 0.0 & 0.0 & 1 & 0 & -- & 10 \\
Truck / default & 0 & 20.0 & 20.0 & 1 & 0 & -- & 5 \\
\multicolumn{8}{l}{\textbf{Qwen3-VL-32B-Thinking / Native GUI Actions (autonomous harness)}} \\
Bolt / easy & 0 & 0.0 & 0.0 & 0 & 0 & -- & 8 \\
Bolt / hard & 0 & 0.0 & 0.0 & 0 & 0 & -- & 8 \\
Color / easy & 0 & 0.0 & 0.0 & 0 & 0 & -- & 10 \\
Color / hard & 0 & 0.0 & 3.8 & 0 & 0 & -- & 10 \\
Maze / easy & 0 & 20.0 & 23.2 & 0 & 0 & -- & 10 \\
Maze / hard & 0 & 0.0 & 6.4 & 0 & 0 & -- & 10 \\
Maze / medium & 0 & 0.0 & 0.0 & 0 & 0 & -- & 10 \\
Nut / easy & 0 & 0.0 & 25.0 & 0 & 0 & -- & 3 \\
Nut / extreme & 0 & 0.0 & 0.0 & 0 & 0 & -- & 3 \\
Nut / hard & 0 & 0.0 & 0.0 & 0 & 0 & -- & 3 \\
Nut / medium & 0 & 0.0 & 0.0 & 0 & 0 & -- & 3 \\
Nut / nightmare & 0 & 0.0 & 0.0 & 0 & 0 & -- & 1 \\
Rush / easy & 0 & 0.0 & 5.4 & 0 & 0 & -- & 10 \\
Rush / hard & 0 & 0.0 & 4.9 & 0 & 0 & -- & 10 \\
Rush / medium & 0 & 0.0 & 0.3 & 0 & 0 & -- & 10 \\
Truck / default & 0 & 0.0 & 18.0 & 0 & 0 & -- & 5 \\
\multicolumn{8}{l}{\textbf{Qwen3-VL-30B-A3B-Thinking / Native GUI Actions (autonomous harness)}} \\
Bolt / easy & 0 & 12.5 & 12.5 & 0 & 0 & -- & 8 \\
Bolt / hard & 0 & 0.0 & 0.0 & 0 & 0 & -- & 8 \\
Color / easy & 0 & 0.0 & 0.0 & 0 & 0 & -- & 10 \\
Color / hard & 0 & 0.0 & 3.8 & 1 & 0 & -- & 10 \\
Maze / easy & 0 & 0.0 & 0.0 & 0 & 0 & -- & 10 \\
Maze / hard & 0 & 0.0 & 1.4 & 1 & 0 & -- & 10 \\
Maze / medium & 0 & 0.0 & 1.4 & 0 & 0 & -- & 10 \\
Nut / easy & 0 & 0.0 & 0.0 & 0 & 0 & -- & 3 \\
Nut / extreme & 0 & 0.0 & 0.0 & 0 & 0 & -- & 3 \\
Nut / hard & 0 & 0.0 & 0.8 & 1 & 0 & -- & 3 \\
Nut / medium & 0 & 0.0 & 0.0 & 0 & 0 & -- & 3 \\
Nut / nightmare & 0 & 0.0 & 4.2 & 0 & 0 & -- & 1 \\
Rush / easy & 0 & 0.0 & 2.5 & 0 & 0 & -- & 10 \\
Rush / hard & 0 & 0.0 & 6.2 & 0 & 0 & -- & 10 \\
Rush / medium & 0 & 0.0 & 0.3 & 0 & 0 & -- & 10 \\
Truck / default & 0 & 0.0 & 2.0 & 0 & 0 & -- & 5 \\
\multicolumn{8}{l}{\textbf{Qwen3-VL-8B-Thinking / Native GUI Actions (autonomous harness)}} \\
Bolt / easy & 0 & 12.5 & 12.5 & 0 & 0 & -- & 8 \\
Bolt / hard & 0 & 0.0 & 0.0 & 0 & 0 & -- & 8 \\
Color / easy & 0 & 0.0 & 1.7 & 0 & 0 & -- & 10 \\
Color / hard & 0 & 0.0 & 2.5 & 0 & 0 & -- & 10 \\
Maze / easy & 0 & 0.0 & 0.0 & 0 & 0 & -- & 10 \\
Maze / hard & 0 & 0.0 & 0.0 & 0 & 0 & -- & 10 \\
Maze / medium & 0 & 0.0 & 7.1 & 0 & 0 & -- & 10 \\
Nut / easy & 0 & 33.3 & 44.4 & 0 & 0 & -- & 3 \\
Nut / extreme & 0 & 0.0 & 0.6 & 0 & 0 & -- & 3 \\
Nut / hard & 0 & 0.0 & 3.0 & 0 & 0 & -- & 3 \\
Nut / medium & 0 & 0.0 & 0.0 & 0 & 0 & -- & 3 \\
Nut / nightmare & 0 & 0.0 & 0.0 & 0 & 0 & -- & 1 \\
Rush / easy & 0 & 0.0 & 0.0 & 0 & 0 & -- & 10 \\
Rush / hard & 0 & 0.0 & 0.0 & 0 & 0 & -- & 10 \\
Rush / medium & 0 & 0.0 & 0.0 & 0 & 0 & -- & 10 \\
Truck / default & 0 & 0.0 & 0.0 & 0 & 0 & -- & 5 \\
\end{longtable}\endgroup

\end{document}